\documentclass[journal,twoside]{IEEEtran}
\usepackage{cite}
\usepackage{amsmath,amssymb,amsfonts}
\usepackage{algorithmic}
\usepackage{graphicx}
\usepackage{textcomp}

\def\BibTeX{{\rm B\kern-.05em{\sc i\kern-.025em b}\kern-.08em
    T\kern-.1667em\lower.7ex\hbox{E}\kern-.125emX}}
\usepackage{graphics} 
\usepackage{epsfig} 
\usepackage{times} 
\usepackage{amsmath} 
\usepackage{amssymb}  
\usepackage{graphicx, bm}
\usepackage{psfrag}
\usepackage{commath}
\usepackage{mathtools}

\usepackage{placeins}

\usepackage{url}
\usepackage{picture}
\usepackage[dvipsnames]{xcolor}
\usepackage{subcaption}
\usepackage{float}
\usepackage{verbatimbox}
\usepackage{booktabs}
\usepackage{listings}
\usepackage{tcolorbox}

\usepackage{tikz}

\usepackage{hyperref}

\newtheorem{itlemma}{Lemma}
\newtheorem{itdefinition}{Definition}
\newtheorem{itproposition}{Proposition}
\newtheorem{itremark}{Remark}
\newtheorem{itcorollary}{Corollary}
\newtheorem{itexample}{Example}

\newenvironment{proposition}{\begin{itproposition}\rm}{\end{itproposition}}

{

  }

\graphicspath{{Images/}}

\begin{document}
\title{UAV-Based Search and Rescue in Avalanches using ARVA: An Extremum Seeking Approach}
\author{Ilario A. Azzollini, Nicola Mimmo, Lorenzo Gentilini, and Lorenzo Marconi
\thanks{The authors are with the Center for Research on Complex Automated Systems (CASY), Department of Electrical, Electronic and Information Engineering (DEI), University of Bologna, Italy (e-mails: ilario.azzollini@unibo.it, nicola.mimmo2@unibo.it, lorenzo.gentilini6@unibo.it, and lorenzo.marconi@unibo.it).}
}

\thispagestyle{empty}
\textit{This work has been submitted to the IEEE for possible publication. Copyright may be transferred without notice, after which this version may no longer be accessible.}

\newpage
\setcounter{page}{0}
\maketitle

\begin{abstract}
This work deals with the problem of localizing a victim buried by an avalanche by means of a drone equipped with an ARVA (Appareil de Recherche de Victimes d'Avalanche) sensor.
The proposed control solution is based on a ``model-free'' extremum seeking strategy which is shown to succeed in steering the drone in a neighborhood of the victim position.
The effectiveness and robustness of the proposed algorithm is tested in Gazebo simulation environment, where a new flight mode and a new controller module have been implemented as an extension of the well-known PX4 open source flight stack.
Finally, to test usability, we present hardware-in-the-loop simulations on a Pixhawk 2 Cube board.

\end{abstract}


\section{Introduction}\label{sec:introduction}
\subsection{The Search \& Rescue avalanche application context}

Nowadays, disasters due to avalanches are even more frequent because of the changing environmental conditions  and the even more marked attitude of people to live extreme mountain experiences, often without the appropriate experience and preparation. Even focusing only on rescuing 	operations on the Italian and Swiss side of the Alps, 2988 people were rescued in alpine accidents due to
	avalanches in the last 15 years with 883 fatalities (source AINEVA and SLF).

Rescue missions in avalanches are characterized by specific peculiarities that make them quite demanding. One of the challenging aspects is the tight constraint imposed on the rescue time.  In fact, survival chances of people buried under the snow decreases rapidly with burial time due to hypothermia. Furthermore, the rescue scenes are typically quite harsh because of  irregular and unstable snow blocks, typically on steep slopes, which make the human intervention complicated, slow and, very often, risky. In fact, it is not rare that the rescuers may trigger a second avalanche event during the S\&R mission. A further critical element is represented by the limited range of sensors that can be used to localize a person buried under meters of snow~\cite{ferrara2015technical}.  One of the most common equipments used in avalanche setting is represented by the ARVA system.

The ARVA equipment has two easily switchable operating modes, which are the \textit{transmitter} and the \textit{receiver} mode. Before starting their activities, experienced skiers switch the worn sensor to the transmitter mode, thus emitting an electromagnetic signal. In case of an accident, companions not buried by the avalanche, or rescuers who reach the disaster area, switch their devices to the receiver mode and start searching the victim by following  well-established ARVA-based search strategy \cite{Azzollini2020Extremum}. The receiver provides information about the electromagnetic field generated by the transmitter sensed at the receiver location. The rescuers are trained to interpret these data to move towards the victim.

The aforementioned tight requirements of the  mission, naturally lead to imagine the development of an aerial robotic platform carrying the ARVA receiver and accomplishing the localization of the ARVA transmitter autonomously. Drones, in fact,  represent a valid support for humans since they can fly autonomously above the snow to find the transmitter location, thus resulting in a faster and safer search. 

The specific application of S\&R in avalanche settings already attracted the interest of the scientific community.  Activities were conducted in the context of the European project \textit{SHERPA} \cite{marconi2012sherpa} where the development of specific robotic technologies  to support professional alpine rescue teams in avalanche scenarios were proposed, and now with the H2020 European project AirBorne \cite{Airborne}, motivating the present work,  whose objective is to develop (at TRL8) a drone equipped with sensor technologies typically used for quick localization of victims. 
In this context, the works \cite{cacace2016control,cacace2016implicit,bevacqua2015mixed} already showed how S\&R operations can greatly benefit from the use of UAVs to survey the environment and collect evidences about the position of people buried under the snow. 

\subsection{State-of-the-art in source seeking algorithms}

The applicative scenario illustrated before frames in a broader research area that is the one referred to as \textit{source seeking control}. In the framework of source seeking, a robotic agent (or a fleet of agents) is able to sense the signal emitted by an omni-directional source located at an unknown position, with the signal strength  having an extremum at the source location. The control problem then consists of processing the signal field measurements, \textit{possibly using a model of it}, to steer the agent (or agents) towards the source. In source seeking the vector field underlying the signal strength is dealt with as the ``map'' to be optimized.
Several approaches have been proposed in literature to solve this class of control problems.  Among the existing ones, a central role for this paper is played by Extremum Seeking (ES).
ES is a real-time model-free optimization approach, which can be used to optimize input-output maps having a \textit{global extremum} (either a minimum or a maximum). It is referred to as model free as no explicit knowledge about this map is required \cite{KrsticBook}. ES could be dated back to 1922 \cite{NesicSurvey} but it has seen a renewed growth in the control community during the last two decades, starting with the proof of local stability in \cite{krstic2000stability} and the extension to semiglobal stability in \cite{Nesic2006}. ES schemes are intrinsically robust and thus appealing for several applications. In particular, talking about control of mobile robots, ES has been used extensively over the last decade for solving source seeking problems \textit{where the model of the source vector field is not available}: the robot has to autonomously find the unknown position of the source, without having any explicit mathematical knowledge of its vector field, therefore by only sensing the source power at the current robot location \cite{zhang2007source}.

Because ES can deal with unknown systems by design, it has been proven to be a powerful tool for steering mobile robots towards a source even in GPS-denied environments \cite{cochran2009source,cochran2009nonholonomic}. Recently, in \cite{Poveda2021}, a class of novel hybrid model-free controllers achieving robust source seeking and obstacle avoidance has been proposed, also in a multi-vehicle scenario. In fact, even with a single agent trying to locate a source, smooth time-invariant feedback controllers based on navigation or barrier functions have been shown to be highly susceptible to arbitrarily small jamming signals that can induce instability in the closed-loop system. Moreover, the problem is not trivial mainly because of the topological obstructions induced by the obstacle.

Besides ES, all the other existing approaches still rely on the intuition that a source localization problem can be formulated as an optimization problem. Inspired by ES, \cite{unicycle2017} shows how the motion limitations arising from using a high-frequency dither signal can be overcome when the typical sinusoidal functions usually employed in ES, already exist in the plant model. In particular, considering a unicycle, they show how ES-like controllers can be developed without adding any external excitation signals, because of the trigonometric nonlinearities of the unicycle model.

Another family of interesting approaches are the \textit{line minimization-based algorithms} \cite{mayhew2008robust}. In these approaches the receiver finds, on a search line, the location of minimum/maximum signal strength. Then, the receiver changes its search path (which belongs to a set of directions that span the whole search space) and iterates the procedure to find the transmitter. 
	
On the other hand, if the radiation pattern is known, the source location could be also obtained via state observers \cite{Salaris2019Online}. Here the main challenge is that of designing sufficiently exciting but also feasible receiver trajectories which ensure the stability of the estimator.  

A problem related to the source seeking is the \textit{boundary tracking problem} \cite{Menon2014Boundary}. In this context it is assumed that the signal iso-strength lines enclose a region of the search domain which contains the source. Then, the receiver may locate the transmitter by exploiting the geometry of these boundary lines.

Finally, \cite{SourceExploration2020} and references therein, deal with bio-inspired optimization techniques. In particular, \cite{SourceExploration2020} presents a planner able to drive an underactuated robot towards the odor source, whose control law is inspired by two prominent behaviors widely observed in biology, namely, chemotaxis and anemotaxis. 

\subsection{Contributions of the work}

This work builds on the previous work \cite{Azzollini2020Extremum}, in which the current ARVA-based S\&R strategy is explained in details, and the modeling and formal investigation of the properties of an ARVA receiver able to sense the complete 3D electromagnetic field are presented. Moreover, in \cite{Azzollini2020Extremum}, we showed how the ARVA output map can be elaborated in order to guarantee continuity, boundedness, and convexity properties and, in this way, how the problem in question can be cast in the ES context. In that paper, the drone dynamics and low-level controller design were ignored.

In this work, we develop an innovative ES-based control solution able to steer an autonomous ARVA-equipped UAV, as close as possible to the victim position. By leveraging on the main properties of ES, the proposed algorithm is not relying on an exact knowledge of the ARVA signal, which is quite uncertain and noisy, but rather on the main features of the ARVA signal in terms of convexity and existence of a unique maximum. In particular, among all the existing ES algorithms, we rely on \cite{scheinker2014extremum}, which is an optimal choice for this application.

The presented control framework is general for solving source seeking problems by means of mobile robots, where ES control can be chosen as reference position generator. In particular, it is shown how the proposed ES scheme can be easily tuned so as to produce smooth position reference signals to be tracked by the robot, taking into account the maximum allowed robot speed and acceleration. Then, we discuss how to guarantee the needed time scale separation between the reference generator and the low-level controller, so as to have the two units working in synergy in a stable way. The proposed low-level controller leverages on the fact that typically, in many ES control schemes like the one we propose, the needed excitation/exploration is provided by having sinusoidal signals to be followed. Thus, we propose an internal model-based controller, leveraging on the fact that the model of the reference signals to be tracked is known.

The specific choices of both the ES algorithm and the low-level controller, are driven by the need of having a complete control scheme which is efficient in terms of computational resources used. In this direction, in order to prove its effectiveness and robustness, the proposed control algorithm has been extensively tested and evaluated through realistic Software-In-The-Loop (SITL) Gazebo simulations. Then, prototyping of the code with Hardware-In-The-Loop (HITL) simulations on a resource constrained microcontroller was performed. In particular, the proposed control algorithm has been implemented as an extension of the open source PX4 flight stack and tested on a Pixhawk 2 Cube board.

Our previous work \cite{Azzollini2020Extremum} was a proof of concept presentation, where we showed how ES could be the tool of choice to solve the ARVA-based S\&R problem after conditioning the ARVA map.
Unlike \cite{Azzollini2020Extremum}, here: (i) we choose the most convenient ES algorithm for searching on a 2D plane with smooth trajectories that could be easily followed by the drone; (ii) we do not assume that the search plane is simply at a certain height with respect to the inertial frame, but rather we take into account the mountain slope; (iii) we develop an internal model-based controller, working in synergy with the ES unit; (iv) we discuss how to add a low-pass filter and tune the parameters so as to guarantee a dynamically feasible trajectory, given the maximum allowed speed and acceleration of the robot; (v) we perform simulations using a realistic drone model and simulation environment, and we also test the code/algorithm performance on a low-cost microcontroller.

This paper is organized as follows. Section \ref{sec:arva} introduces the ARVA system and describes the problem at hand. Section \ref{sec:solution} presents the problem solution discussing the role of the different control units one by one. In Section \ref{sec:practical} we discuss the implementation details and tuning choices, and we show the practical results. Finally, the conclusions are presented in Section \ref{sec:conclusions}.

\subsection{Notation}

$I_n \in \mathbb{R}^{n \times n}$ is used to denote the $n$-dimensional identity matrix, while $0_{n \times m}$ denotes a $n \times m$ matrix of zeros.
With $SO(3)$ it is denoted the \textit{special orthogonal group} of 3D rotation matrices, i.e. $SO(3) = \{ R \in \mathbb{R}^{3 \times 3} : R^\top R = R R^\top = I_3, \text{det}R = 1\}$, while $SE(3) = \{ H \in \mathbb{R}^{4 \times 4} : H = \begin{bmatrix}
	R & o \\
	0_{1 \times 3} & 1
	\end{bmatrix},\text{ s.t. } R \in SO(3), \; o \in \mathbb{R}^3\}$.
	For a differentiable function $g$, its gradient is denoted by $\nabla g$.

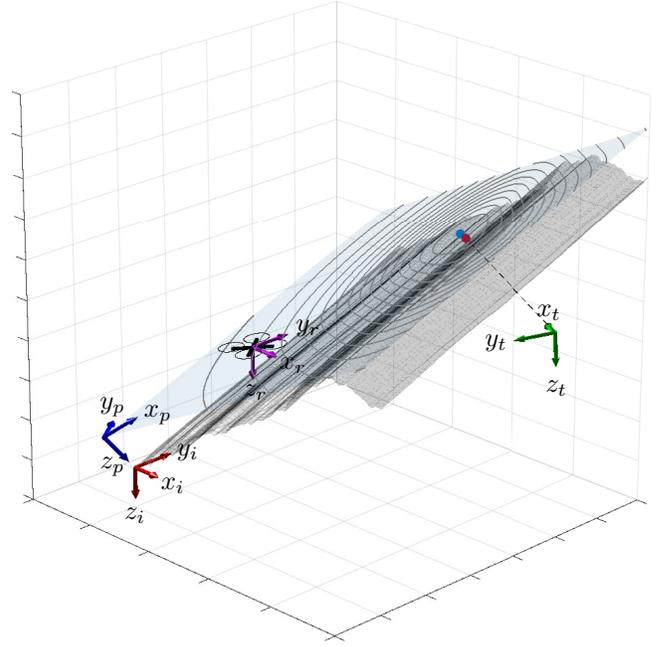
\begin{figure}
	\centering
	\begin{tikzpicture}
		\node (image) at (0,0) {\includegraphics[trim = 300 0 250 0, clip=true, width=\columnwidth]{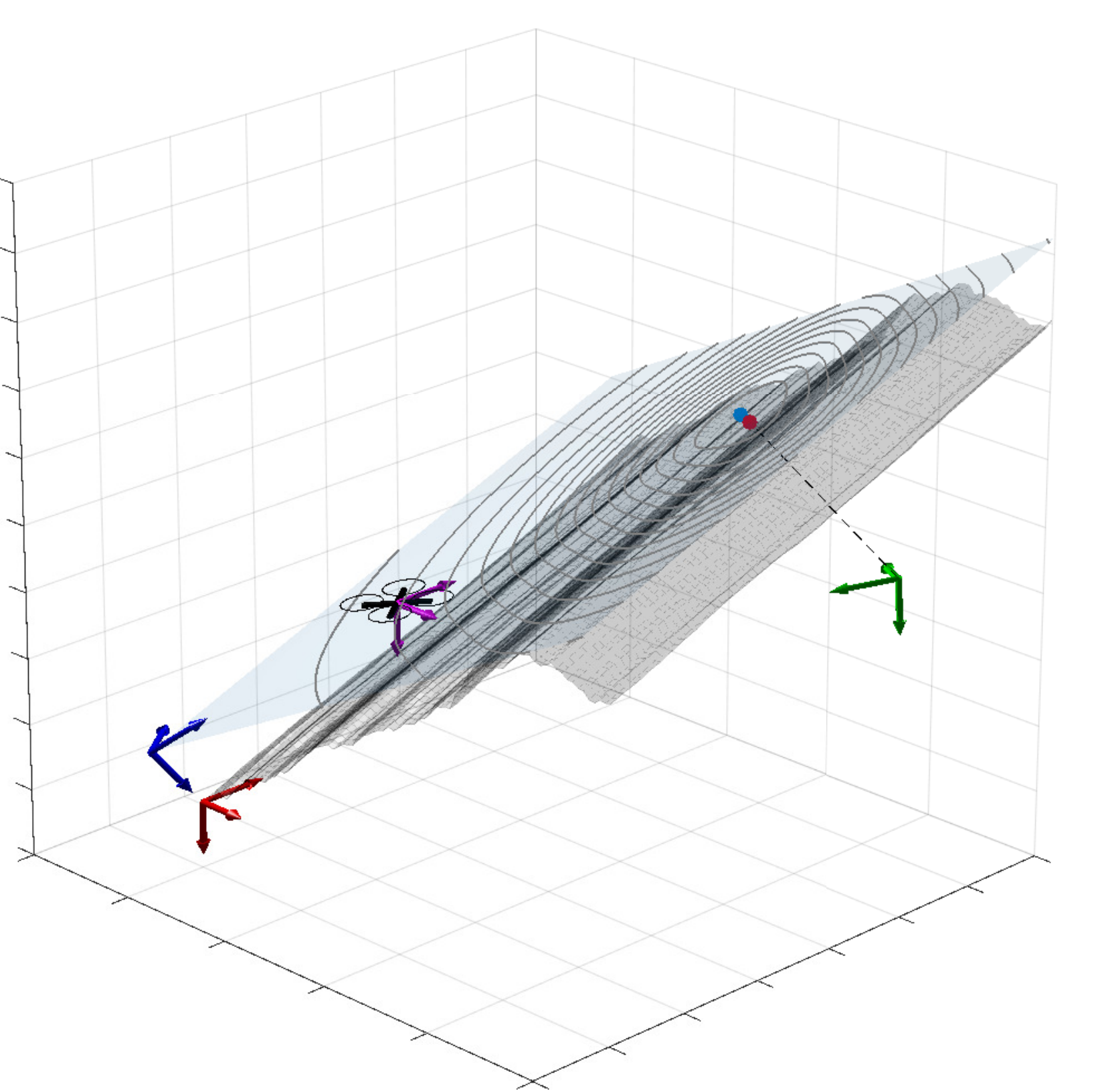}};
		\node[] (xi) at (-2.3,-2.3) {$x_i$};
		\node[] (yi) at (-2.1,-1.85) {$y_i$};
		\node[] (zi) at (-2.8,-2.7) {$z_i$};
		\node[] (xp) at (-2.5,-1.35) {$x_p$};
		\node[] (yp) at (-3.1,-1.25) {$y_p$};
		\node[] (zp) at (-3.1,-2.05) {$z_p$};
		\node[] (xd) at (-0.7,-0.7) {$x_r$};
		\node[] (yd) at (-0.5,-0.2) {$y_r$};
		\node[] (zd) at (-1.2,-1.05) {$z_r$};
		\node[] (xt) at (2.7,0) {$x_t$};
		\node[] (yt) at (2.0,-0.4) {$y_t$};
		\node[] (zt) at (2.8,-1) {$z_t$};		
		\end{tikzpicture}
	\caption{Reference frames: one to identify the inertial space, one to describe the search plane, and two to denote the transmitter and the receiver pose, respectively.}
	\label{fig:RefFrames}
\end{figure}

In this manuscript, four Cartesian coordinate frames are defined (see Figure \ref{fig:RefFrames}):
${\cal F}_i = \left({\rm O}_i,\, {x}_i,\, {y}_i,\, {z}_i\right)$ denotes the right-handed static inertial frame, with origin ${\rm O}_i$, with the axis ${x}_i$ oriented toward geographic north, ${z}_i$ oriented opposite to the local gravity vector and ${y}_i$ oriented to create a right-handed frame (i.e. North-East-Down, or simply NED), while  ${\cal F}_t = \left(O_t,\, {x}_t,\, {y}_t,\, {z}_t\right)$ and ${\cal F}_r = \left({\rm O}_r,\, {x}_r,\, {y}_r,\, {z}_r\right)$ are the right-handed frames associated to the static transmitter worn by the victim and to the receiver installed on the moving drone, respectively. Moreover, ${\cal F}_p = \left(O_p,\, {x}_p,\, {y}_p,\, {z}_p\right)$ defines the reference frame for the search plane description. In more details, $z_p$ is orthogonal to the search plane whereas $O_p$ lives on the search plane. For the sake of simplicity we assume that the body frame, attached to the centre of gravity of the drone, coincides with ${\cal F}_r$.
 The positions of ${\rm O}_r$  and ${\rm O}_t$ relative to ${\rm O}_p$ are indicated by the vectors ${ p}_r \in \mathbb{R}^3$ and ${ p}_t  \in \mathbb{R}^3$, respectively. Given that ${\rm O}_t$ and ${\rm O}_p$ are static reference frames, ${ p}_t$ is a constant. The position of  ${\rm O}_r$ relative to ${\rm O}_t$ is indicated by the vector $ p \in \mathbb{R}^3$, with $p=p_r-p_t$. Throughout the paper, we shall use the superscripts $i$, $p$, $t$ and $r$ on the left of the vectors $ p$, ${ p}_t$, ${ p}_r$ to denote the representation of the previous vectors in the reference frames ${\cal F}_i$, ${\cal F}_p$, ${\cal F}_t$ and ${\cal F}_r$, respectively (for instance, $\prescript{p}{}{p}$ denotes a representation of $ p$ in ${\cal F}_p$).
 Finally $e_3 = [0 \ 0 \ 1]^\top$ and $S(x)$, with $x = [x_1 \ x_2 \ x_3]^\top \in \mathbb{R}^3$, denotes the skew-symmetric matrix
\begin{equation*}
	S(x) =
	\begin{bmatrix}
		0 & -x_3 & x_2 \\
		x_3 & 0 & -x_1 \\
		-x_2 & x_1 & 0
	\end{bmatrix}.
\end{equation*} 
The orientation of a moving frame ${\cal F}_\#$ with respect to the inertial frame ${\cal F}_i$ can be expressed by means of the sequence of extrinsic elementary rotations denoted by roll $\phi_\#$ (about the $x$-axis), pitch $\theta_\#$ (about the $y$-axis), and yaw $\psi_\#$ (about the $z$-axis). We use the formalism in \cite[Section 2.4.2]{siciliano2010robotics} to compute the related overall rotation matrix $\prescript{i}{}{R}_{\#}(\psi_\#,\theta_\#,\phi_\#) \in SO(3)$ (from the $\#$-frame to the inertial frame).
 
\section{The ARVA system}\label{sec:arva}
In this section, we first go through the main physical principles of the ARVA system with the final goal to derive a model of the signal vector field. Then, we present the related search strategy and we describe the problem we want to solve.

\subsection{Modeling}
\label{sec:Modeling}

The transceivers commercially available have two operating modes, namely they can work as receivers or as transmitters, with a manual switch used to commute between the two. In \textit{transmission} mode the ARVA generates a magnetic field that is modeled as a dipole aligned with the ${x}_t$ axis of ${\cal F}_t$. The electromagnetic vector field, described in $\mathcal{F}_p$, is indicated by $\prescript{p}{} h \in \mathbb{R}^3$. By letting $ \prescript{p}{}p= \prescript{p}{}p_r - \prescript{p}{}p_t = [x \ y \ z]^\top$, it turns out that a mathematical model of the magnetic vector field is given by \cite{pinies2006fast}
\begin{equation}
\label{eq:ARVAMagnField}
\prescript{p}{}h(\prescript{p}{}p,\prescript{p}{}R_{t}) = \dfrac{1}{4\pi \|\prescript{p}{}p\|^5} \,{\rm A}(\prescript{p}{}p) \prescript{p}{}R_{t}e_1
\end{equation}
where      
\begin{equation*}
{\rm A}(\prescript{p}{}p) := \left[\begin{array}{ccc}
2x^2-y^2-z^2 & 3xy & 3xz\\
3xy &2y^2-x^2-z^2 &  3yz\\
3xz & 3yz & 2z^2-x^2-y^2 \\
\end{array}\right]
\end{equation*}
and $e_1 = [1 \ 0 \ 0]^\top$. The flux lines described by the previous model are symmetric with respect to the transmitter $x_t$ axis. 
The intensity of the magnetic field  can be then obtained by the previous relation as (see \cite{pinies2006fast})
\begin{equation}
\label{eq:Hmodideal}
\|\prescript{p}{} h\| = \dfrac{1}{4\pi \|\prescript{p}{} p \|^3}\sqrt{1+3\dfrac{\prescript{p}{}p^\top M \prescript{p}{}p }{\|\prescript{p}{} p\|^2}}
\end{equation}
where $M = \prescript{t}{}R_p^\top  e_1  e_1^\top \prescript{t}{}R_p \ge 0$ with minimum and maximum singular values given by $\underline{\sigma}(M) = 0$ and $\overline{\sigma}(M) = 1$, respectively. It turns out that $\|\prescript{p}{} h\|$ is radially unbounded with $1/\| ^pp \|$, namely the intensity of the magnetic field is infinity when $p_r=p_t$. 
 Furthermore, \eqref{eq:Hmodideal} can be exploited to compute the iso-power lines, that are also symmetric with respect to the transmitter $x_t$ axis.
 
 In the context of this S\&R application, what is relevant is the projection of the flux and iso-power fields onto the so-called {\em search plane}, conveniently identified as the $x_p y_p$-plane, (see Figure \ref{fig:RefFrames}). The search plane is the plane on which the drone is required to operate, and it is chosen to be parallel to the snow surface, at a safe distance from the ground. This distance should be kept as small as possible, with a minimum imposed by the irregularities of the terrain and the presence of possible rescuers on the avalanche scene. The overall distance between the victim-transmitter and the chosen search plane is denoted by $d_t$. Therefore, we can simply write the position of the transmitter with respect to the search plane frame ${\cal F}_p$, as $\prescript{p}{}{p}_{t} = [t_x \ t_y \ d_t]^\top$. Ideally, we would like to drive the drone on the geometric projection of $O_t$ on the search plane, that is simply given by $\prescript{p}{}{p}_{t / {\rm proj}} = [t_x \ t_y \ 0]^\top$, as this is clearly the closest admissible position to the transmitter location.
 
 However, the projection of the iso-power field onto the search plane is affected by the distance $d_t$ and by the rotation matrix $\prescript{p}{}R_{t}$, parameterizing the orientation of the transmitter with respect to the search plane frame. In fact, the optimal position $p^\star$, corresponding to the maximum intensity of the magnetic field that we can sense on the plane, usually differs from $\prescript{}{}{p}_{t / {\rm proj}}$.
 As a first example, in Figure \ref{fig:RefFrames}, the distance $d_t$ is indicated by the dashed line, the geometric projection position $\prescript{}{}{p}_{t / {\rm proj}}$ is given by the red dot, while the optimal position $p^\star$ is indicated with the blue dot.
 
\begin{figure}[h!]
	\centering
\begin{subfigure}[t]{0.45\columnwidth}
	\centering
	\includegraphics[width=\textwidth]{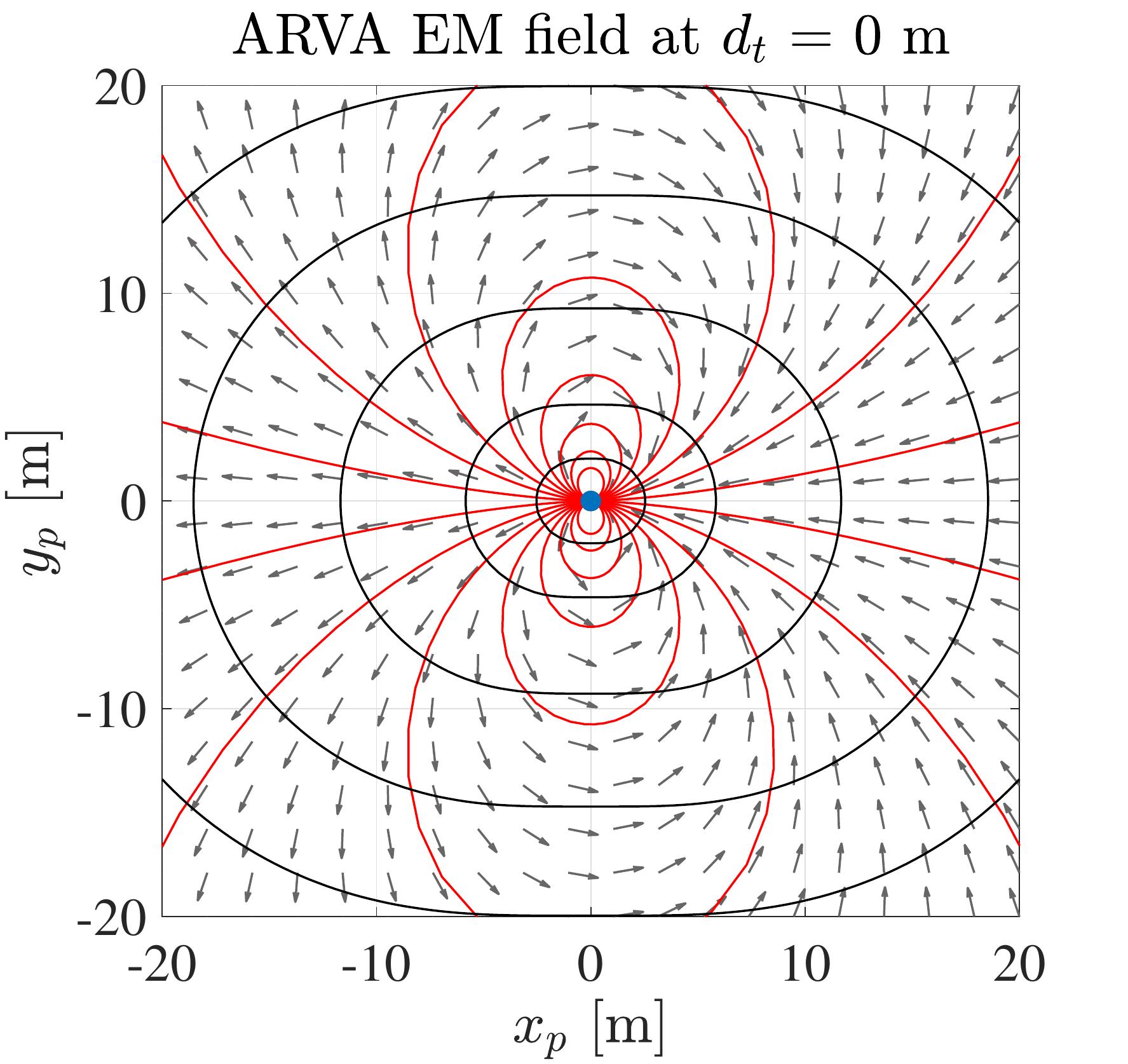}
	\caption{$\prescript{p}{}R_{t} = I_3$, $d_t = 0$.}
	\label{fig:MagnDipole1}
\end{subfigure}
\begin{subfigure}[t]{0.45\columnwidth}
	\centering
	\includegraphics[width=\textwidth]{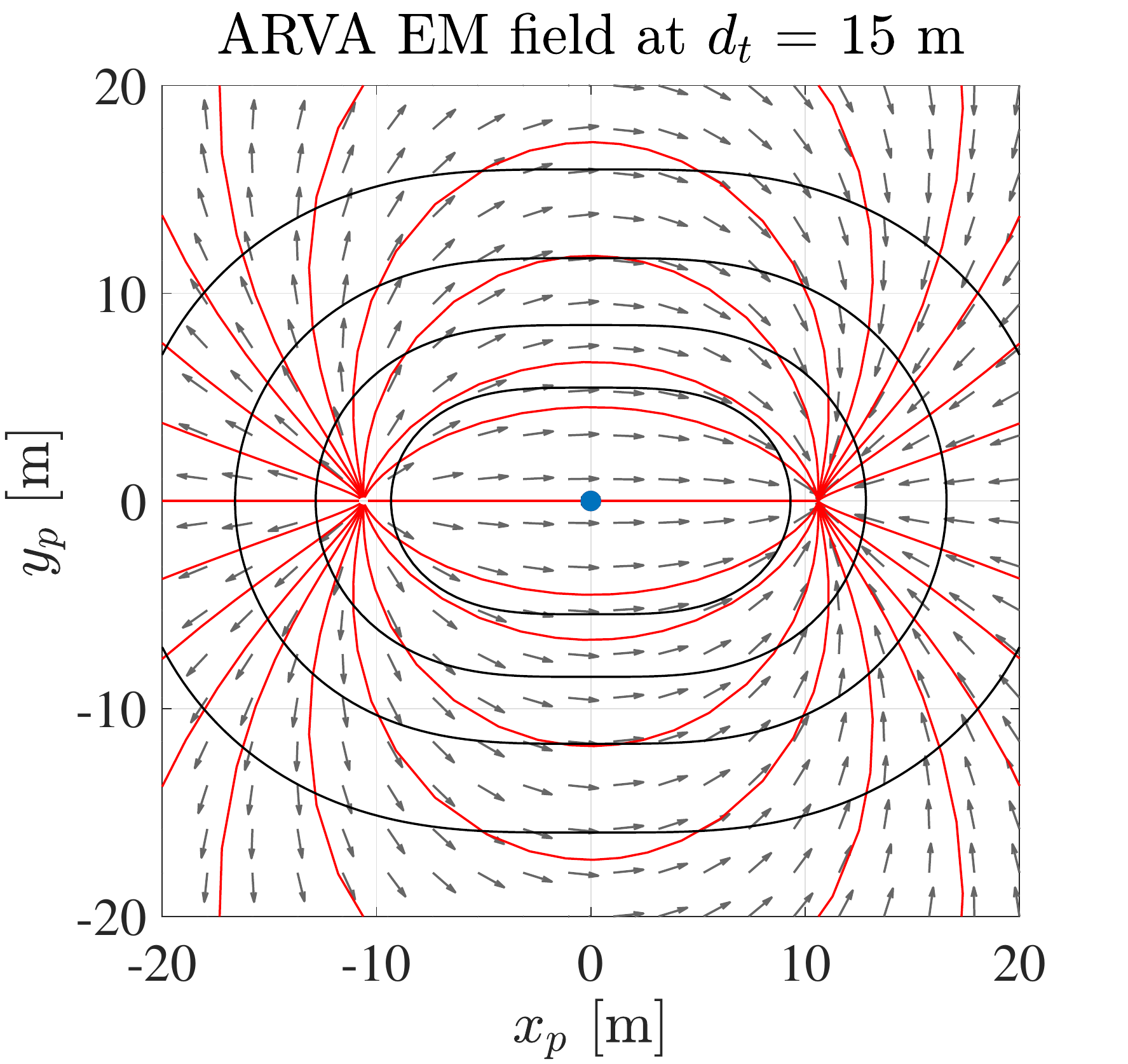}
	\caption{$\prescript{p}{}R_{t} = I_3$, $d_t = 15$.}
	\label{fig:MagnDipole2}
\end{subfigure}

\begin{subfigure}[t]{0.45\columnwidth}
	\centering
	\includegraphics[width=\textwidth]{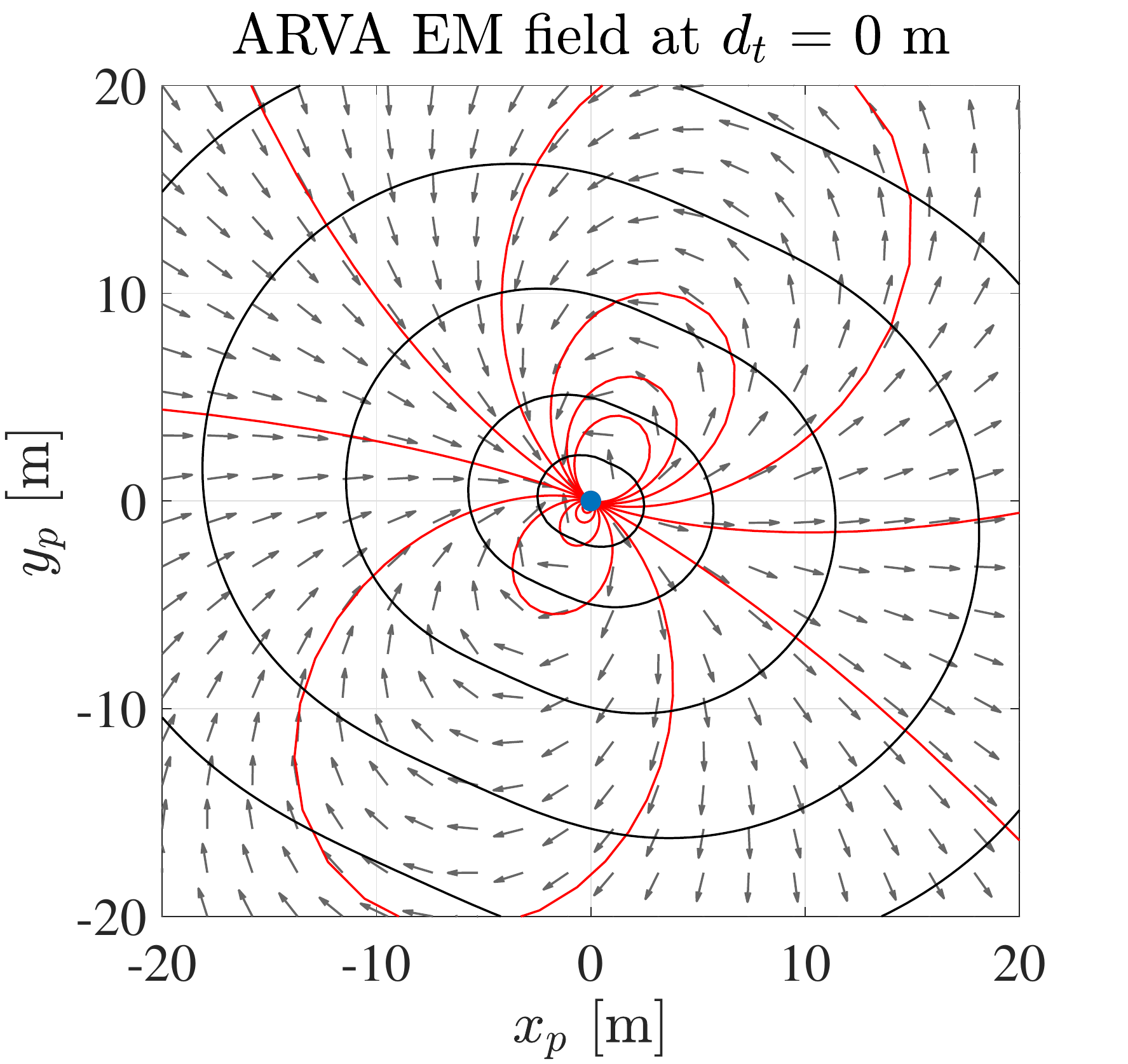}
	\caption{$\prescript{p}{}R_{t} \neq I_3$, $d_t = 0$.}
	\label{fig:MagnDipole1rot}
\end{subfigure}
\begin{subfigure}[t]{0.45\columnwidth}
	\centering
	\includegraphics[width=\textwidth]{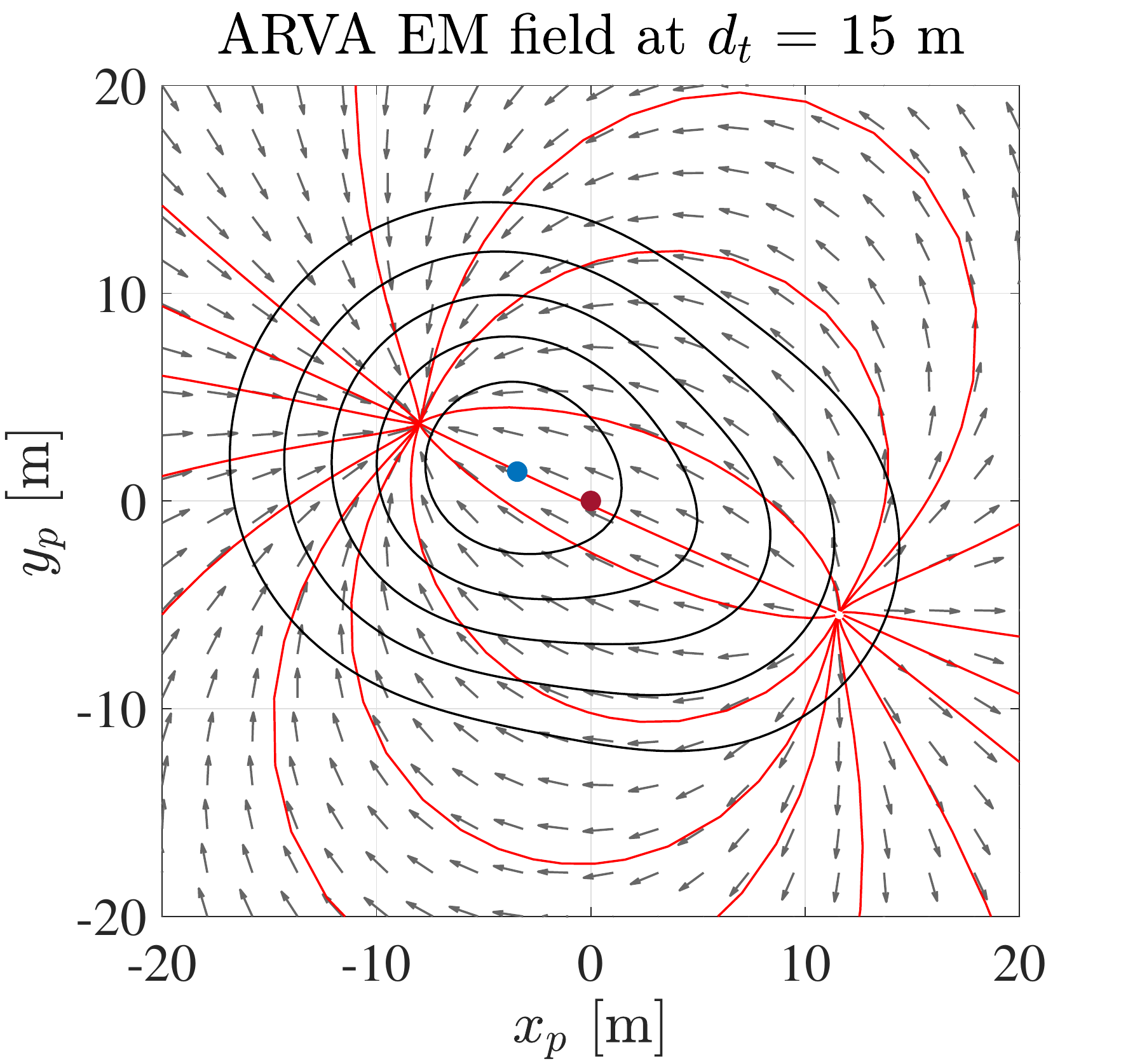}
	\caption{$\prescript{p}{}R_{t} \neq I_3$, $d_t = 15$.}
	\label{fig:MagnDipole2rot}
\end{subfigure}
\caption{ARVA flux lines (in red), EM vector field (arrows), and iso-power lines (in black). The transmitter is located at $\prescript{p}{}{p}_{t} = [0 \ 0 \ d_t]^\top$, its geometric projection onto the search plane is the red dot $\prescript{p}{}{p}_{t / {\rm proj}} = [0 \ 0 \ 0]^\top$, while the ARVA EM field maximizer $p^\star$ is the blue dot.}
\label{fig:fluxlinesIso}
\end{figure}

 In order to further understand this aspect, we can look at the EM field restricted to the search plane in Figure \ref{fig:fluxlinesIso}. In particular, both the flux lines (in red) and the iso-power lines (in black) are depicted for different instances  of $d_t$ and $\prescript{p}{}R_{t}$, assuming for simplicity that the transmitter location is $\prescript{p}{}{p}_{t} = [0 \ 0 \ d_t]^\top$. As a matter of fact, $\prescript{p}{}{p}^{\star} = \prescript{p}{}{p}_{t / {\rm proj}}$ only in the (unlikely to happen) scenarios in which either $d_t=0$ or $\prescript{p}{}R_{t} = I_3$. In all the other cases, $\prescript{}{}{p}^{\star}$ will only be in a neighborhood of $\prescript{}{}{p}_{t / {\rm proj}}$.

The ARVA signal is received through three antennas directed along the \textit{receiver} frame axes ${x}_r$, ${y}_r$ and  ${z}_r$, namely along the longitudinal, lateral and  vertical direction of the sensor case. The magnetic field sensed at the receiver location, denoted by $\prescript{p}{}h_m$, is given by
\begin{equation}
\label{eq:ARVAmeas}
\prescript{p}{}h_m(\prescript{p}{}p,\prescript{p}{}R_{t},w) = \prescript{p}{}h(\prescript{p}{}p,\prescript{p}{}R_{t}) +  \prescript{p}{}w(t)
\end{equation}       
where $\prescript{p}{}w:\mathbb{R} \mapsto \mathbb{R}^3$ indicates the ElectroMagnetic Interferences (EMI) expressed in the search plane frame.
There are two sources of EMI, the drone and the environment. Small drones are commonly actuated through electromagnetic brushless motors governed by logic units constituted by switches powered by LiPo batteries. The whole electrical power distribution chain is prone to the emission of EM noises which are sensed by the ARVA receiver. Fortunately, since these interferences can be investigated in dedicated EM testing facilities, the drones under development in \cite{Airborne} will be equipped with special shields that minimise the on-board generated EMI. On the contrary, the environment clearly cannot be modified to reduce the EM noise. Usually, the environmental electromagnetic field is affected by the presence of power lines, funicular railways, etc. These effects are suitably modeled through signals, namely $\prescript{p}{}{w}(t)$, whose amplitude is bounded and quasi-constant on the avalanche search area, \textit{i.e.} there exists $\overline{w} > 0$ such that $\|\prescript{p}{}{w}(t)\|_\infty \leq \overline{w}$. Finally, it is interesting to notice that the power density of the ideal dipole goes to infinity at the transmitter location and is a strictly decreasing function of $\|\prescript{p}{}{p}\|$. This, beside the boundedness of  $\prescript{p}{}{w}$, leads to
\begin{equation}
	\begin{array}{c}
		\lim\limits_{\|\prescript{p}{}p\|\to \infty} \|\prescript{p}{}h_m\|_{\infty} = \bar{w},\quad
		\lim\limits_{\|\prescript{p}{}p\|\to 0} \|\prescript{p}{}h_m\|_{\infty} = \infty
	\end{array}
\end{equation}
which will be exploited in Section \ref{sec:Measurement}.

\subsection{Existing search strategies and problem description}

	\begin{figure}[]
		\centering
		\includegraphics[width=0.8\columnwidth]{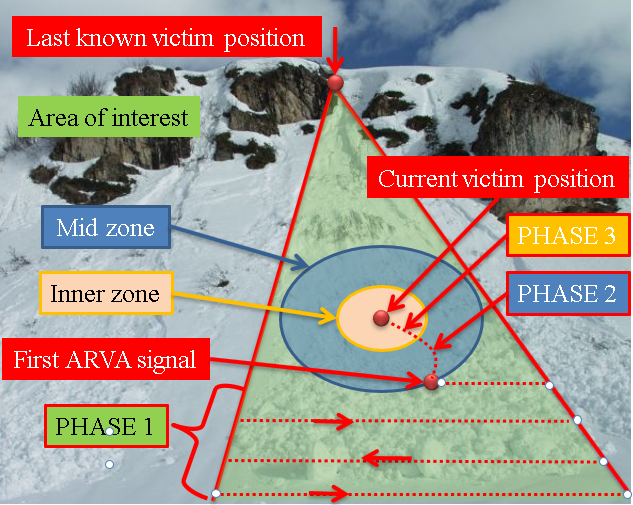}
		\caption{Rescue scene: area of interest and search phases.}
		\label{fig:SearchTech}
	\end{figure}
	
The ARVA-based search strategy is graphically sketched in Figure \ref{fig:SearchTech}.
After the definition of the so-called \textit{area of interest}, which is a triangular area starting from the last known victim position and including the avalanche front, the search is divided in three subsequent phases. Starting from the bottom of the area of interest, the \textit{first search phase} consists in following straight parallel lines with an offset of 15-20 meters, with the goal of finding a valid ARVA signal. When sufficiently close to the transmitter (typically around 50 meters), a valid ARVA signal is measured, and the \textit{second search phase} starts. The ARVA receiver displays the EM vector field in terms of magnitude and direction, which actually corresponds to the tangent to the EM field flux line at the operator location. The rescuers are trained to follow the flux line to approach the victim. The \textit{third search phase} begins when the sensed EM field is sufficiently strong, namely the ARVA receiver is sufficiently close to the victim and automatically changes its output modality, providing only the modulus of the EM field at the operator location. The automatic change of modality is thought to inform the rescuers that the flux line approach is no more efficient, and therefore they start searching by iteratively applying a two-step gradient search strategy (which consists in finding maximum EM intensity along orthogonal directions).

Since the first phase does not hide any particular control challenges (it consists of controlling the drone along pre-established trajectories), we mainly focus on the second and the third phases by assuming the availability of a valid ARVA signal.
In fact, we merge the second and third search phases in a single one based on the processing of the ARVA EM intensity and ignoring the geometry of the flux lines typically considered in the second search phase.  The flux line following strategy, in fact, has several drawbacks, which could be better understood by looking at Figure \ref{fig:fluxlinesIso}: (i) the search path can be unpredictably long as it depends on the initial position of the receiver with respect to the transmitter; (ii) the search path depends on the initial receiver attitude with respect to the transmitter (the rescuers could follow the flux lines counterclockwise or clockwise thus leading to different search paths); (iii) because of the EMI noise, the measure of the EM directions can be particularly deteriorated.

The envisaged scenario is thus the following. First of all, the area of interest as well as the drone search plane are defined by the rescue team (see Figure \ref{fig:RefFrames} and Figure \ref{fig:SearchTech}), based on the last known victim position, and the slope of the terrain with respect to the inertial frame. The drone will then take off and reach the search plane, and will start autonomously performing the first search phase to find a valid ARVA signal. At this point, the valid signal indicates that the victim is located at approximately a 50 meters distance from the current position of the drone. Now, in a practice-inspired design philosophy, a search strategy based on a gradient-like policy could lead to the unique extremum on the chosen search plane.
Thus, this paper aims at designing an automatic control law based on ES, only driven by the intensity of the ARVA EM field, which steers the drone as close as possible to the victim location.

\section{An Extremum Seeking-based solution}\label{sec:solution}
\begin{figure*}[t!]
\centering
\includegraphics[scale=0.9]{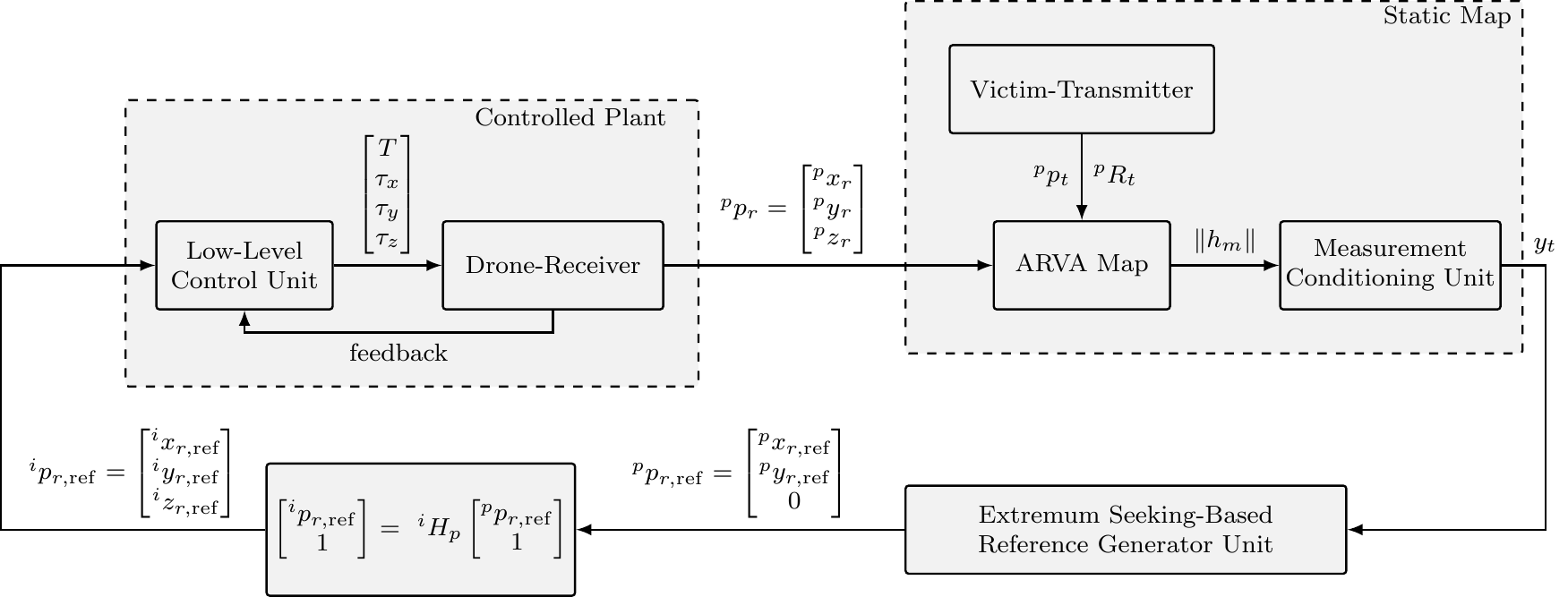}

\caption{Overall control scheme.}
\label{fig:ControlArchitecture}

\end{figure*}

The overall control scheme is sketched in Figure \ref{fig:ControlArchitecture},  where we can distinguish three main units, which are the {\em measurement conditioning unit}, the {\em ES-based reference generator unit} and the {\em low-level control unit}. As already mentioned, we want the ARVA receiver to have only one output modality, that is, to directly provide the intensity of the EM field at the operator location. Therefore, given as inputs the drone-receiver position $\prescript{p}{}{p}_r$, the victim-transmitter position $\prescript{p}{}{p}_t$, and the transmitter orientation $\prescript{p}{}R_{t}$, the {\em ARVA map} block gives as output $\|h_m\|$, which is the intensity of the measurement \eqref{eq:ARVAmeas}.

\subsection{Measurement conditioning unit}
\label{sec:Measurement}

Being the maximum intensity of \eqref{eq:ARVAmeas} equal to infinity, any gradient-based algorithm would face with issues in the proximity of the victim. This criticism motivates the following manipulation. The measurement conditioning unit statically processes the ARVA  intensity measurement $\|h_m\|$ to create a new intensity map, denoted by  $y_t$, that  is continuous and bounded for any $\prescript{p}{}p \in \mathbb{R}^3$, and has a global minimum equal to zero. Specifically, the conditioned measurement $y_t$ is generated as  
\begin{equation}
\label{eq:output}
y_t(\prescript{p}{}p,\prescript{p}{}R_{t},\prescript{p}{}w) := {1 \over \sqrt[3]{\|h_m\|}}.
\end{equation}
Simple computations show that $y_t$ can be approximated by $y_t =  \prescript{p}{}h_{\text{n}} + \nu_t$ in which 
\begin{equation}
\label{eq:nominal}
\prescript{p}{}h_{\text{n}}(\prescript{p}{}p,\prescript{p}{}R_{t}) := \dfrac{(4\pi)^{1/3} \|\prescript{p}{}p\|}{\sqrt[6]{1+3\dfrac{\prescript{p}{}p^\top M \prescript{p}{}p }{\|\prescript{p}{}p\|^2}}}
\end{equation}
is the nominal conditioned intensity and 
\begin{equation}
\label{eq:noise}
\nu_t(\prescript{p}{}p,\prescript{p}{}R_{t},t) = \Xi_A(\prescript{p}{}p,\prescript{p}{}R_{t}) \|\prescript{p}{}p\|^3
 \prescript{p}{}w(t) 
\end{equation}
is the equivalent additive noise in which  $\Xi_A(\prescript{p}{}p,\prescript{p}{}R_{t}) \in \mathbb{R}^3$ is a bounded function.

The new output map \eqref{eq:output} shows some key properties. First, it is well defined because for any $\prescript{p}{}p \in \mathbb{R}^3$  
\begin{equation}
0\le\dfrac{\prescript{p}{}p^\top M \prescript{p}{}p }{\|\prescript{p}{}p\|^2} \le 1.
\end{equation}

In addition, for any fixed $\prescript{p}{}R_{t} \in SO(3)$  the functions $\prescript{p}{}h_\text{n}(\,\cdot\, ,\prescript{p}{}R_{t})$, $\nu_t(\,\cdot\,,\prescript{p}{}R_{t},t)$ both have a global minimum at $\prescript{p}{}p = 0$ and are strictly increasing.
Furthermore, let the Noise-to-Signal Ratio (NSR) to be defined as 
\begin{equation}
{\text{NSR}}(\prescript{p}{}p,\prescript{p}{}R_t) := \dfrac{ \|y_t(\prescript{p}{}p,\prescript{p}{}R_{t},t)\|_{\infty}-|\prescript{p}{}h_\text{n}(\prescript{p}{}p,\prescript{p}{}R_{t})| }{\|y_t(\prescript{p}{}p,\prescript{p}{}R_{t},t)\|_{\infty}}.
\end{equation}
This modified ratio belongs to the compact domain $[0,\,1]$ and, in particular, for any $\prescript{p}{}R_t \in SO(3)$
\begin{equation}
\lim\limits_{\prescript{p}{}p\to 0} {\text{NSR}}(\prescript{p}{}p,\prescript{p}{}R_t) = 0, \quad 
\lim\limits_{\prescript{p}{}p\to \infty} {\text{NSR}}(\prescript{p}{}p,\prescript{p}{}R_t) = 1
\end{equation} 
meaning that at the origin $\prescript{p}{}p= 0$ the output is not affected by noise whereas for $\prescript{p}{}p\to \infty$ the nominal signal is annihilated by the noise.

In conclusion, the conditioned map \eqref{eq:output} can be optimized by means of any (approximate) gradient-based optimization technique. Because of the presence of the noise and the model uncertainty, we do not want to rely on an exact knowledge of the model of $y_t$, but rather on its convexity property and existence of a minimum. For this reasons, ES is chosen, resulting in a robust and practically implementable control algorithm.

\subsection{ES-based reference generator unit}
\label{sec:Esunit}

The conditioned output map $y_t$, constrained on the search plane, has a unique extremum (which is a minimum). The optimal position corresponding to the minimum of $y_t$, clearly coincides with the position relative to the maximum intensity of the ARVA map without conditioning. Therefore the goal is still that of driving the receiver to the optimal point $\prescript{p}{}p^\star$. In fact, $\prescript{p}{}p^\star$ now corresponds to the minimum of $y_t$ restricted to the search plane, and is still the blue dot we saw as an example in Figures \ref{fig:RefFrames} and \ref{fig:fluxlinesIso}

The reference generator unit and the low-level control unit need to work (and to be designed) in synergy. The ES-based unit processes the conditioned ARVA map $y_t$, and its role is ideally that of performing a real-time optimization on the search plane, thus driving the drone-receiver position $\prescript{p}{}p_r$ towards the optimal point on the search plane $\prescript{p}{}p^\star$. As we do not have direct control on the optimization variable $\prescript{p}{}p_r$, ES here plays the role of a reference position generator for the drone. A (low-level) reference tracking controller needs to be designed so as to drive the drone position to the generated reference position $\prescript{p}{}p_{r,{\rm ref}}$, that will be the output of the ES. In this scenario, in order to guarantee proper functioning of the proposed scheme, the controlled plant needs to work on a faster time scale with respect to the ES unit \cite{KrsticBook,NesicSurvey,krstic2000stability}. In fact, we do not have a static input-output map to be optimized, but rather the steady-state input-output map of a dynamical system, with input $\prescript{p}{}p_{r,{\rm ref}}$ and output $y_t$. Requiring the controlled plant to be way faster than the reference generator, the ES design can be made by considering the controlled plant as if it was a static map, so assuming to have direct control on $\prescript{p}{}p_r$.

Now, let us define the components of $\prescript{p}{}p^\star$ and $\prescript{p}{}p_{r, {\rm ref}}$ as $\prescript{p}{}p^\star = [\prescript{p}{}x^\star \; \prescript{p}{}y^\star \; 0]^\top$ and $\prescript{p}{}p_{r, {\rm ref}} = [\prescript{p}{}x_{r, {\rm ref}} \; \prescript{p}{}y_{r, {\rm ref}} \; \prescript{p}{}z_{r, {\rm ref}}]^\top$. The following proposition, adapted from \cite{scheinker2014extremum}, presents the chosen 2-dimensional ES algorithm.

\begin{proposition}\label{proposition:ESalgorithm}
	For any $\delta > 0$, by a sufficiently large choice of $\kappa \alpha$, the point $(\prescript{p}{}x^\star, \prescript{p}{}y^\star)$ is $(1/\omega)$-Semiglobally Practically Uniformly Ultimately Bounded with ultimate bound $\delta$, relative to the system $(\prescript{p}{}x_{r, {\rm ref}}(t), \prescript{p}{}y_{r, {\rm ref}}(t))$:
	\begin{equation}\label{eq:ChosenES}
	\begin{split}
		\prescript{p}{}{\dot{x}}_{r,{\rm ref}} & = \sqrt{\alpha \omega} \cos (\omega t + \kappa y_t) \qquad \prescript{p}{}{x}_{r,{\rm ref}}(0) = \prescript{p}{}{x}_{r}(0) \\ 
		\prescript{p}{}{\dot{y}}_{r,{\rm ref}} &= \sqrt{\alpha \omega} \sin (\omega t + \kappa y_t) \qquad \prescript{p}{}{y}_{r,{\rm ref}}(0) = \prescript{p}{}{y}_{r}(0). \\ 
	\end{split}
\end{equation}
\end{proposition}

The reference signal $\prescript{p}{}p_{r, {\rm ref}}$ is obviously completed with $\prescript{p}{}{z}_{r,{\rm ref}} = 0$, as we want the drone-receiver to always move on the search plane.

This ES unit processes the conditioned ARVA intensity $y_t$ and generates the reference signals for the drone, expressed in the search plane frame $\mathcal{F}_p$. In particular, \eqref{eq:ChosenES} achieves ES in a practical way, meaning that $(\prescript{p}{}x_{r, {\rm ref}}, \prescript{p}{}y_{r, {\rm ref}})$ converge to a neighborhood of $(\prescript{p}{}x^\star, \prescript{p}{}y^\star)$, which can be made arbitrarily small by properly choosing the design parameters $\omega$, $\kappa$, and $\alpha$. Moreover, this result is semiglobal as there exists a certain domain of attraction around $(\prescript{p}{}x^\star, \prescript{p}{}y^\star)$, such that, if we start inside this region we can solve the problem. This domain of attraction can be arbitrarily enlarged by properly choosing the design parameters, at the expense of slowing down the convergence speed.

The presented ES scheme works as follows. System \eqref{eq:ChosenES} is evolving in circular trajectories on the $x_py_p$-plane, where the parameter $\omega$ plays the role of the oscillation frequency. In particular, at steady-state the system geometric path is given by a circumference of radius $\sqrt{\alpha \omega}$ around the optimum $(\prescript{p}{}x^\star, \prescript{p}{}y^\star)$, parameterized in time with the frequency $\omega$. In fact, it can be proven \cite{scheinker2014extremum} that the trajectories of \eqref{eq:ChosenES} uniformly converge to the trajectories $(\prescript{p}{}{\bar{x}},\prescript{p}{}{\bar{y}})$ of the so-called ``average'' system
\begin{equation}\label{eq:averagedynamics}
	\begin{bmatrix}
		\prescript{p}{}{\dot{\bar{x}}} \\
		\prescript{p}{}{\dot{\bar{y}}} 
	\end{bmatrix}
	= - \frac{\kappa \alpha}{2} (\nabla y_t(\prescript{p}{}{\bar{x}},\prescript{p}{}{\bar{y}}))^{\top},
	\qquad
	\begin{bmatrix}
		\prescript{p}{}{\bar{x}}(0) \\
		\prescript{p}{}{\bar{y}}(0)
	\end{bmatrix}
	=
	\begin{bmatrix}
		\prescript{p}{}x_{r, {\rm ref}}(0) \\
		\prescript{p}{}y_{r, {\rm ref}}(0)
	\end{bmatrix},
\end{equation}
which exhibits a stable gradient-flow dynamics, with adaptation gain $\kappa \alpha$. Therefore, for any ultimate bound $\delta > 0$, by choosing arbitrarily large values of $\kappa \alpha$ we may ultimately bound $(\prescript{p}{}{\bar{x}},\prescript{p}{}{\bar{y}})$ within a $\delta$ neighborhood of $(\prescript{p}{}x^\star, \prescript{p}{}y^\star)$.

Averaging technique \cite[Section 10.4]{khalil2002nonlinear} is always used to analyze ES schemes and shows how the system is evolving, on average, in the gradient-descent direction to seek the minimizer. In particular, in the original coordinates, it is the centre of the circular trajectory which is approaching $(\prescript{p}{}x^\star, \prescript{p}{}y^\star)$ in an approximate gradient-descent fashion. The centre of the circular trajectory can be therefore regarded as the current estimate of the optimum. By moving in circular trajectories, we are basically exploring a neighborhood of the current position (i.e. the parameter estimate), to check in which direction the sensed function value is decreasing.

Among all the existing algorithms, this one was chosen for three main reasons. First of all, the generated reference trajectory is guaranteed to be smooth enough to be followed by our UAV.
Moreover, this algorithm is called ``bounded update rates'' ES, as the magnitude of the velocity (which corresponds to the update rate of the estimate) can be a priori chosen as $\lVert v_r \rVert  = \sqrt{\alpha \omega}$. Finally, this algorithm is easy to be discretized and implemented, and it is also light to be executed on a microcontroller.

The parameters to be tuned are the positive scalars $\alpha$, $\omega$, and $\kappa$. By looking at Proposition 1, they are not difficult to be tuned in general, when we simply have a static map to be optimized. However, in our specific application, we have to consider the noise and the drone. In particular, being the ARVA map noisy, we want to choose $\kappa$ small. In fact, the reference signal could be significantly deteriorated by $\kappa y_t$ in \eqref{eq:ChosenES}. As a consequence, in order to have the learning rate $\kappa \alpha$ sufficiently large (as required by Proposition 1), we have to take a large $\alpha$. Now, a tradeoff needs to be considered, as we would also like to take $\omega$ big, following the fact that $1/\omega$ plays the role of $\epsilon$ in the $(\epsilon,\delta)$-SPUUB stability result of Proposition 1 \cite{scheinker2014extremum}. On the other hand, $\sqrt{\alpha \omega}$ corresponds to the maximum speed of the drone, as well as to the radius of the circumference which the drone describes on the search plane. In practice, we will see that a good choice is that of taking $\alpha$ very large and $\omega$ quite small such that the maximum allowed speed is respected. In this way, being the frequency of oscillations not very high, while having a big circumference radius, the reference signal will be gentle enough to be followed by the drone in near-hovering condition at all times.

Moreover, we introduce the low-pass filter
\begin{equation}\label{eq:alpha_filter}
			\dot{\alpha} = -\dfrac{1}{\lambda} \alpha + \dfrac{1}{\lambda} \alpha_{\rm max} \hspace{1.75cm} \alpha(0) = 0 \\
\end{equation}
resulting in $\alpha$ in \eqref{eq:ChosenES} that approaches the chosen $\alpha_{\rm max}$ in an arbitrary amount of time (starting from zero). This gives us full authority to impose also the preferred maximum acceleration, so as to ensure a dynamically feasible trajectory. In conclusion, given a maximum velocity and a maximum acceleration, a dynamically feasible reference trajectory can always be guaranteed for our UAV, by an appropriate choice of the parameters in \eqref{eq:ChosenES}, \eqref{eq:alpha_filter}.

As a final part of this unit, we define the homogeneous transformation matrix $\prescript{i}{}H_p$ such that
\begin{equation}
	\begin{bmatrix}
		^ix_{r, \text{ref}} \\
		^iy_{r, \text{ref}} \\
		^iz_{r, \text{ref}} \\
		1
	\end{bmatrix} = \ 
	^iH_p
	\begin{bmatrix}
		^px_{r, \text{ref}} \\
		^py_{r, \text{ref}} \\
		0 \\
		1
	\end{bmatrix}.
\end{equation}
so as to design a low-level controller for the UAV in the more convenient inertial frame $\mathcal{F}_i$.

\subsection{Low-level Control Unit}

The low-level controller aims at driving the drone position to the generated reference position as fast as possible, so as to ensure the time scale separation needed for the proper functioning of the ES unit.  A model-based approach based on linearization is followed in the design of the controller as presented next. It must be stressed, though, that any  favorite reference tracking controller for UAV can be taken, provided that the time scale separation requirement is fulfilled. However, we choose to develop linearization-based controllers as we only have a reference position to track, without any specific reference for the attitude. This means we can operate the UAV in near-hovering conditions at all times, without requiring a more complicated and computationally heavier controller. This intuition will be supported by the experimental results, where good tracking performance will be achieved, even in hardware-in-the-loop simulations while using a low-cost microcontroller.
 
As the references are generated by the designed ES optimizer, their model is perfectly known. In particular, we know that \textit{at steady-state} the drone-receiver will be asked to track a biased sinusoidal signal of known frequency $\omega$ for each of the position components $\prescript{i}{}{x}_r$, $\prescript{i}{}{y}_r$, and $\prescript{i}{}{z}_r$. This follows by the fact that we are generating a circumference as the geometric path reference on the 2D search plane, which is mapped to an ellipse in the 3D space.


We start by considering the nonlinear dynamical model of vertical take-off and landing aerial vehicles by means of the well-known Newton-Euler rigid body equations \cite{hua2013introduction}
\begin{subequations}\label{eq:UAV_NL_model}
\begin{align}
		\label{eq:a} M \prescript{i}{}{\ddot{p}}_r &= -T \prescript{i}{}{R}_r e_3 + M g e_3 \\ 
		\label{eq:b} \prescript{i}{}{\dot{R}}_r &= \prescript{i}{}{R}_r S(\prescript{r}{}{\omega}_r) \\
		\label{eq:c} J \prescript{r}{}{\dot{\omega}}_r &= -S(\prescript{r}{}{\omega}_r) J \prescript{r}{}{\omega}_r + \tau
\end{align}
\end{subequations}
in which $M > 0 \in \mathbb{R}$ and $J = {\rm diag}(J_{x},J_{y},J_z) \in \mathbb{R}^{3 \times 3}$ are the UAV mass and inertia matrix, respectively, $\prescript{i}{}p_{r} = [\prescript{i}{}x_r \ \prescript{i}{}y_r \ \prescript{i}{}{z}_r]^\top$ denotes the position of the centre of gravity of the system expressed in the inertial frame, $\prescript{r}{}{\omega}_r = [\prescript{r}{}{\omega}_x, \prescript{r}{}{\omega}_y, \prescript{r}{}{\omega}_z]^\top \in \mathbb{R}^3$ is the angular speed expressed in the drone-receiver frame, $\prescript{i}{}{R}_{r}(\psi_r,\theta_r,\phi_r) \in SO(3)$ is the rotation matrix from the drone-receiver frame to the inertial frame, while $T>0 \in \mathbb{R}$ and $\tau = [\tau_x \ \tau_y \ \tau_z]^\top \in \mathbb{R}^3$ are the thrust force and vector of torques, respectively. Finally, recall the definitions of $e_3$ and the skew-symmetric matrix $S(\prescript{r}{}{\omega}_r)$ from the Notation, as well as the fact that also the $z_r$ axis of the drone-receiver frame points downwards.

The system \eqref{eq:UAV_NL_model} is then linearized around the \textit{hovering} equilibrium point, namely
\begin{equation}
	\begin{cases}
		\prescript{i}{}{p}_r^{\star} &= \prescript{i}{}{p}_r^{\rm hov} \\
		\prescript{i}{}{\dot{p}}_r^{\star} &= 0_{3 \times 1} \\
		\prescript{i}{}{R}_r^{\star} &= I_3 \\
		\prescript{r}{}{\omega}_r^{\star} &= 0_{3 \times 1}
	\end{cases} \qquad
	\begin{cases}
		T^{\star} &= Mg \\
		\tau^{\star} &= 0_{3 \times 1}
	\end{cases}
\end{equation}
where $\prescript{i}{}{p}_r^{\rm hov}$ is any arbitrary hovering position. In a gain scheduling fashion, we linearize online taking $\prescript{i}{}{p}_r^{\rm hov} = \prescript{i}{}{p}_{r, {\rm ref}}$. It is well-known that the linearized system results in four independent systems, namely the \textit{roll}, the \textit{pitch}, the \textit{yaw}, and the \textit{vertical} dynamics. On the four subsystems, the corresponding four inputs are then designed as indicated in the following.

\textit{Yaw dynamics and control:} The yaw dynamics results in
\begin{equation}\label{eq:YawDyn}
\begin{bmatrix}
			\dot{\psi}_r \\
			\ddot{\psi}_r
\end{bmatrix} =
\begin{bmatrix}
	0 & 1 \\
	0 & 0
\end{bmatrix}
\underbrace{\begin{bmatrix}
	\psi_r \\
	\dot{\psi}_r
\end{bmatrix}}_{X_y} +
\begin{bmatrix}
	0 \\
	\dfrac{1}{J_z}
\end{bmatrix} \tau_z
\end{equation}
and we simply choose to stabilize it with a state-feedback controller of the form
	$\tau_z = -K_y X_y$, with $K_y \in \mathbb{R}^{1 \times 2}$ to be designed such that the resulting closed-loop system is Hurwitz.
This choice corresponds to asking the controller to keep the yaw angle to zero (we assume that the yaw is equal to zero when starting the search). As already mentioned, this simple control choice can be made as we do not need to point the receiver towards the transmitter (or in general, towards the direction we are going), but we only need to move the receiver on the search plane, regardless of the ``heading'' angle.
	
\textit{Roll dynamics and control:} We start by defining the error $\prescript{i}{}e_y := \prescript{i}{}{y}_r - \prescript{i}{}{y}_{r,{\rm ref}}$. The roll dynamics result in
\begin{equation}\label{eq:RollDyn}
\begin{bmatrix}
			\prescript{i}{}{\dot{y}}_r \\
			\prescript{i}{}{\dot{v}}_y \\
			\dot{\phi}_r \\
			\ddot{\phi}_r
\end{bmatrix} =
\begin{bmatrix}
	0 & 1 & 0 & 0\\
	0 & 0 & g & 0 \\
	0 & 0 & 0 & 1 \\
	0 & 0 & 0 & 0
\end{bmatrix}
\underbrace{\begin{bmatrix}
			\prescript{i}{}e_y \\
			\prescript{i}{}{v}_y \\
			\phi_r \\
			\dot{\phi}_r
\end{bmatrix}}_{X_r} +
\begin{bmatrix}
	0 \\
	0 \\
	0 \\
	\dfrac{1}{J_x}
\end{bmatrix} \tau_x.
\end{equation}

The goal of the control loop is to let the lateral drone position $\prescript{i}{}{y}_r$ tracking the reference $\prescript{i}{}{y}_{r,{\rm ref}}$ computed by the ES-based reference generator system. As previously discussed this reference signal (as well as $\prescript{i}{}{x}_{r,{\rm ref}}$ and $\prescript{i}{}{z}_{r,{\rm ref}}$) is a biased sinusoidal signal of unknown amplitude but known frequency $\omega$, coming from \eqref{eq:ChosenES}. Thus, in order to drive the regulation error $\prescript{i}{}e_y$ to zero, we consider an internal model-based regulator (see \cite[Chapter 4]{IsidoriBook2017}) of the form
\begin{equation}
\begin{split}
	\dot{\eta}_r &=
	\underbrace{\begin{bmatrix}
		0 & 1 & 0 \\
		0 & 0 & 1 \\
		0 & -\omega^2 & 0
	\end{bmatrix}}_{\Phi}
	\eta_r +
	\underbrace{\begin{bmatrix}
		0 \\
		0 \\
		1
	\end{bmatrix}}_{G}
	\prescript{i}{}e_y \\
	\tau_x &= - K_r X_r - K_{\eta_r} \eta_r
\end{split}
\end{equation}
with $K_r$ and $K_{\eta_r}$ to be designed such that the resulting closed-loop system is Hurwitz.

\textit{Pitch dynamics and control:} As before, define the error $\prescript{i}{}e_x := \prescript{i}{}{x}_r - \prescript{i}{}{x}_{r,{\rm ref}}$. The pitch dynamics are given by
\begin{equation}\label{eq:PitchDyn}
\begin{bmatrix}
			\prescript{i}{}{\dot{x}}_r \\
			\prescript{i}{}{\dot{v}}_x \\
			\dot{\theta}_r \\
			\ddot{\theta}_r
\end{bmatrix} =
\begin{bmatrix}
	0 & 1 & 0 & 0\\
	0 & 0 & -g & 0 \\
	0 & 0 & 0 & 1 \\
	0 & 0 & 0 & 0
\end{bmatrix}
\underbrace{\begin{bmatrix}
			\prescript{i}{}e_x \\
			\prescript{i}{}{v}_x \\
			\theta_r \\
			\dot{\theta}_r
\end{bmatrix}}_{X_p} +
\begin{bmatrix}
	0 \\
	0 \\
	0 \\
	\dfrac{1}{J_y}
\end{bmatrix} \tau_y.
\end{equation}
Analogously as before, the internal model-based regulator is
\begin{equation}
\begin{split}
	\dot{\eta}_p &= \Phi \eta_p + G \prescript{i}{}e_x \\
	\tau_y &= - K_p X_p - K_{\eta_p} \eta_p
\end{split}
\end{equation}
with $K_p$ and $K_{\eta_p}$ to be designed such that the resulting closed-loop system is Hurwitz.


\textit{Vertical dynamics and control:} Define the error $\prescript{i}{}e_z := \prescript{i}{}{z}_r - \prescript{i}{}{z}_{r,{\rm ref}}$. The vertical dynamics are given by
\begin{equation}\label{eq:VertDyn}
\begin{bmatrix}
			\prescript{i}{}{\dot{z}}_r \\
			\prescript{i}{}{\dot{v}}_z
\end{bmatrix} =
\begin{bmatrix}
	0 & 1 \\
	0 & 0
\end{bmatrix}
\underbrace{\begin{bmatrix}
		\prescript{i}{}{e}_z \\
		\prescript{i}{}{v}_z
\end{bmatrix}}_{X_v} +
\begin{bmatrix}
	0 \\
	-\dfrac{1}{M}
\end{bmatrix} (T - T^\star)
\end{equation}
The internal model-based regulator is given by
\begin{equation}
\begin{split}
	\dot{\eta}_v &= \Phi \eta_v + G \prescript{i}{}e_z \\
	T &= - K_v X_v - K_{\eta_v} \eta_v
\end{split}
\end{equation}
with $K_v$ and $K_{\eta_v}$ to be designed such that the resulting closed-loop system is Hurwitz.
\section{Implementation and Results}\label{sec:practical}
In order to promote flexibility and encourage usage as well as further improvements, the presented algorithm has been implemented as an extension of the open-source PX4 flight software \cite{meier2015px4}. Simulations are carried out exploiting the Gazebo-based simulation environment RotorS \cite{furrer2016rotors} along with the provided model of the 3DR Iris quadrotor, properly modified to carry the latest available ARVA receiver plugin \cite{Cacace2021ARVA}.  The adopted Iris model is also equipped with an essential sensor suite composed of an Inertial Measurement Unit (IMU) and a GPS.
Exploiting the PX4 firmware modular structure, we implemented our algorithm combining a new ``extremum seeking'' PX4 module jointly with a new flight mode called ``search'', so that we could easily switch to this mode when a first ARVA signal is found and the developed algorithm will autonomously start working. Moreover, slight modifications to the current multicopter control loop have been implemented to admit the new regulator. The already provided PX4 Extended Kalman Filter (EKF) module has been used.

To evaluate the performances of the proposed algorithm, we present two different simulation scenarios that comprise Software-In-The-Loop (SITL) and Hardware-In-The-Loop (HITL) simulations. In order to make the two presented simulation results comparable, the same set-up is used.
In particular, the drone-receiver is initially located at $\prescript{i}{}{p}_r(0) = [\prescript{i}{}{x}_r = 0, \prescript{i}{}{y}_r = 0, \prescript{i}{}{z}_r = -6]^\top$, corresponding to the position in space where the first ARVA signal has been detected after the first search phase. For simplicity, we also take this starting point as coincident with the origin of the search plane frame.

In order to be compliant with the maximum range of action of commercial ARVA sensors, the victim location has been randomly chosen to be initially approximately $50$ meters far from the initial receiver position.
We have a distance between the victim-transmitter and the search plane equal to $d_t=15$ m. Moreover, the transmitter orientation with respect to the search plane frame $\prescript{p}{}{R}_t$ has been numerically computed to obtain the worst case scenario in terms of distance between the optimal position on the search plane $p^\star$, and the geometric projection of the victim position on the search plane $p_{t / {\rm proj}}$. Therefore we are in a case similar to that of Figure \ref{fig:MagnDipole2rot}. This scenario is of great practical importance because, as victims are usually buried at a distance between 0.5 and 10 meters from the snow plane, the performed simulations really represent a worst case scenario. In fact, the rescuers in charge of the last part of the rescue operations involving digging and finding the victim, are well trained and able to quickly save the victim if they are given an estimate $p^\star$ which is located in a 10 meters radius from the unknown $p_{t / {\rm proj}}$ (they perform the digging on the orthogonal direction with respect to the search plane, and thus with respect to the snow plane).

\begin{table}[b]
	\begin{center}
		\captionsetup{width=0.7\columnwidth}
		\caption{Simulation parameters.}\label{tab:SimParam}
		\begin{tabular}{lc}\hline
			\hline
			{\scriptsize Victim $x$-position in inertial coordinates} & {\scriptsize $24.0866$}       \\
			{\scriptsize Victim $y$-position in inertial coordinates} & {\scriptsize $34.0866$}       \\
			{\scriptsize Victim $z$-position in inertial coordinates} & {\scriptsize $-16.8773$}       \\
			{\scriptsize Roll angle (transmitter to inertial) $\phi_t$} & {\scriptsize $0$}       \\
			{\scriptsize Pitch angle (transmitter to inertial) $\theta_t$} & {\scriptsize $0.1745$}       \\
			{\scriptsize Yaw angle (transmitter to inertial) $\psi_t$} & {\scriptsize $2.7052$}      \\
			{\scriptsize $^iO_p$} & {\scriptsize $[0 \ 0 \ -6.1268]^\top$}       \\
			{\scriptsize Roll angle (transmitter to search plane) $\phi_p$} & {\scriptsize $0$}       \\
			{\scriptsize Pitch angle (transmitter to search plane) $\theta_p$} & {\scriptsize $0.6162$}       \\
			{\scriptsize Yaw angle (transmitter to search plane) $\psi_p$} & {\scriptsize $0.7854$} \\
			\hline
		\end{tabular}
	\end{center}
	\vspace{-5pt}
\end{table}
\begin{figure*}[t!]
	\centering
	\begin{subfigure}[b]{0.31\textwidth}	
		\centering
		\includegraphics[clip = true, width = \textwidth, height = 0.8\textwidth]{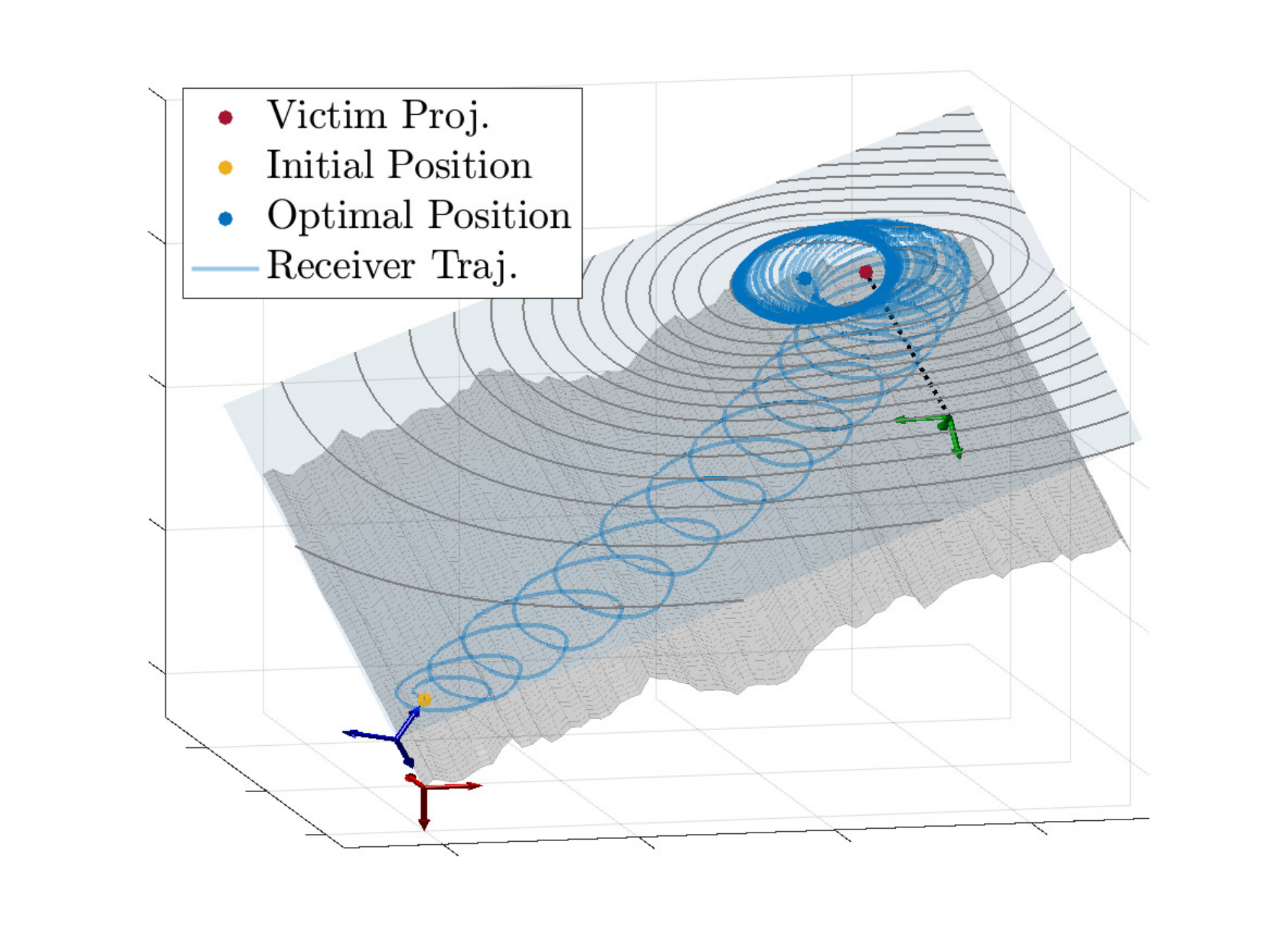}
		\caption{Drone trajectory: 3D view.}
		\label{fig:sim_results(a)}
		\vspace{2pt}	
		\centering
		\includegraphics[clip = true, width = \textwidth, height = 0.8\textwidth]{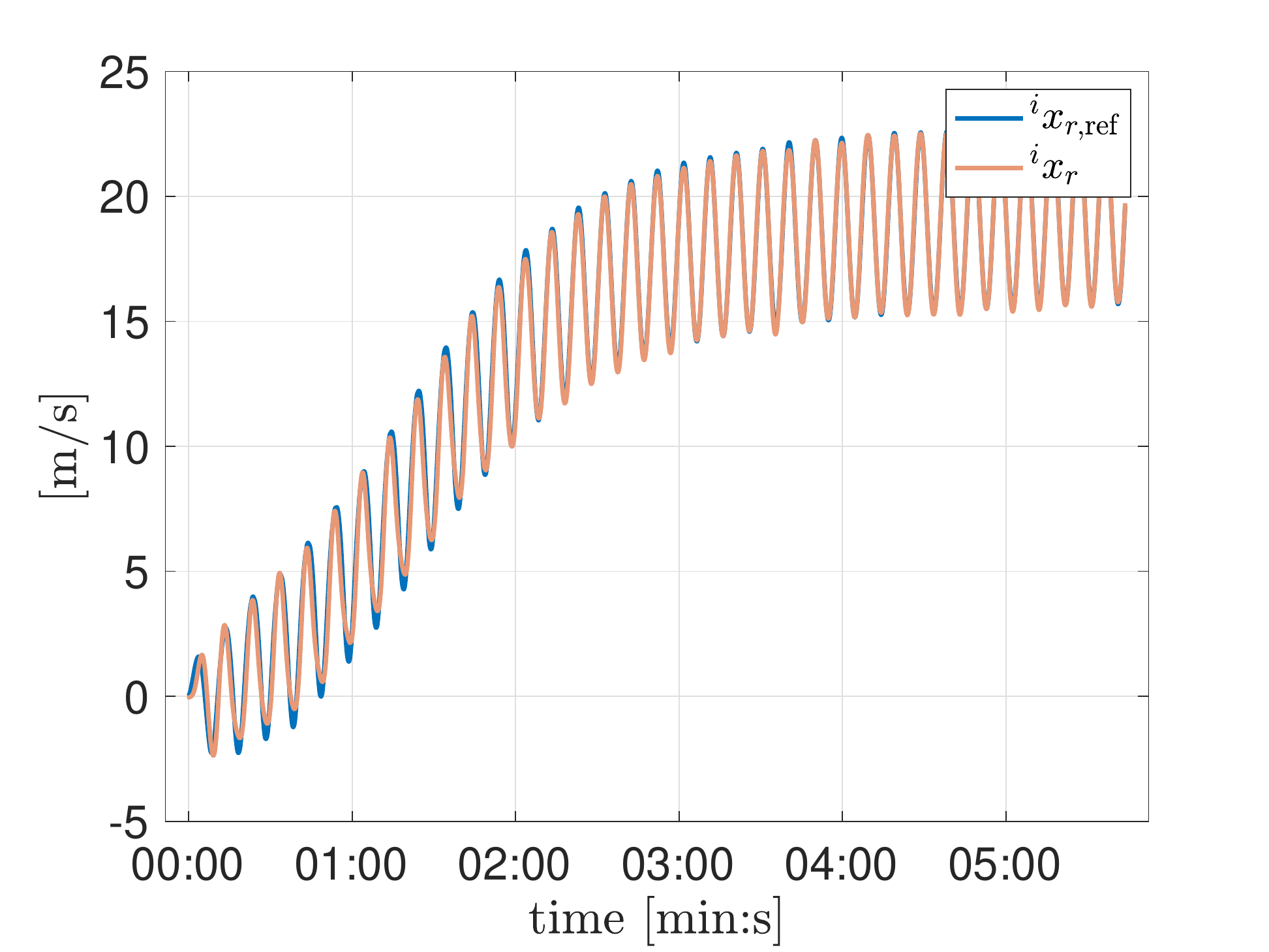}
		\caption{Reference tracking: $x$-component.}
		\label{fig:sim_results(b)}
		\vspace{2pt}
		\centering
		\includegraphics[clip = true, width = \textwidth, height = 0.8\textwidth]{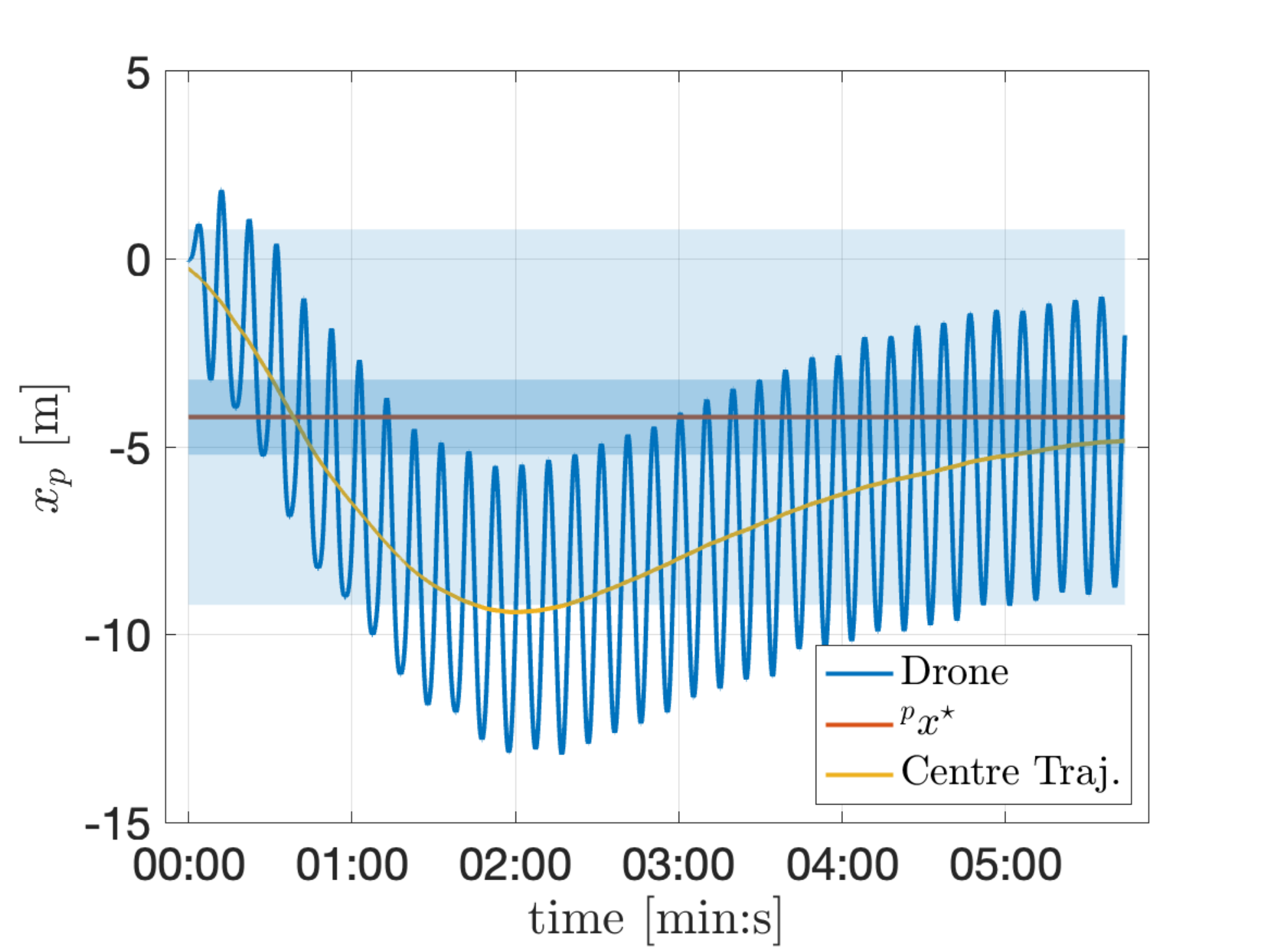}
		\caption{Optimum seeking: $x$-component.}
		\label{fig:sim_results(c)}
	\end{subfigure}
	\hspace{0.01\textwidth}
	\begin{subfigure}[b]{0.31\textwidth}
		\centering
		\includegraphics[clip = true, width = \textwidth, height = 0.8\textwidth]{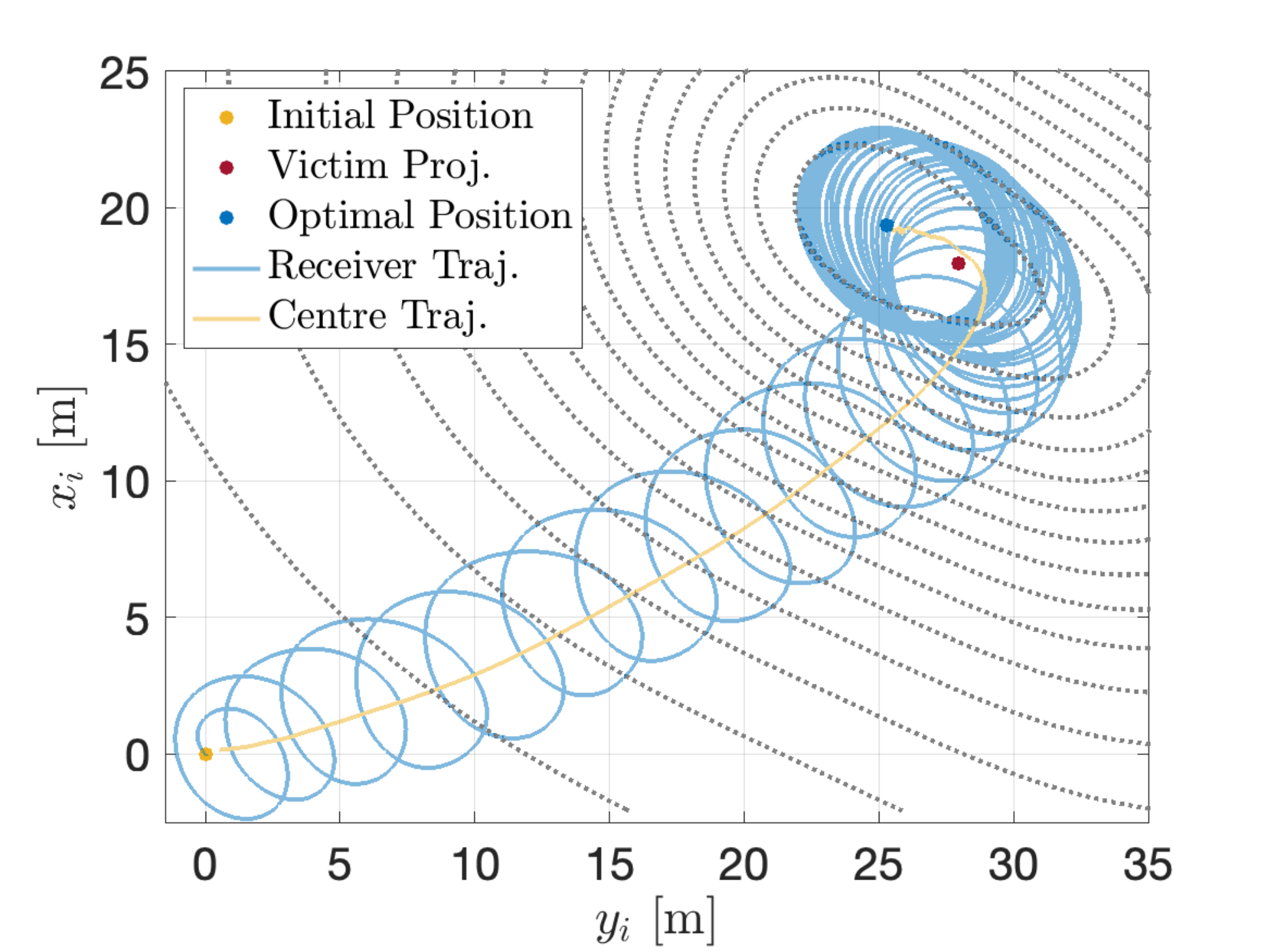}
		\caption{Drone trajectory: $x_iy_i$-plane view.}
		\label{fig:sim_results(d)}
		\vspace{2pt}
		\centering
		\includegraphics[clip = true, width = \textwidth, height = 0.8\textwidth]{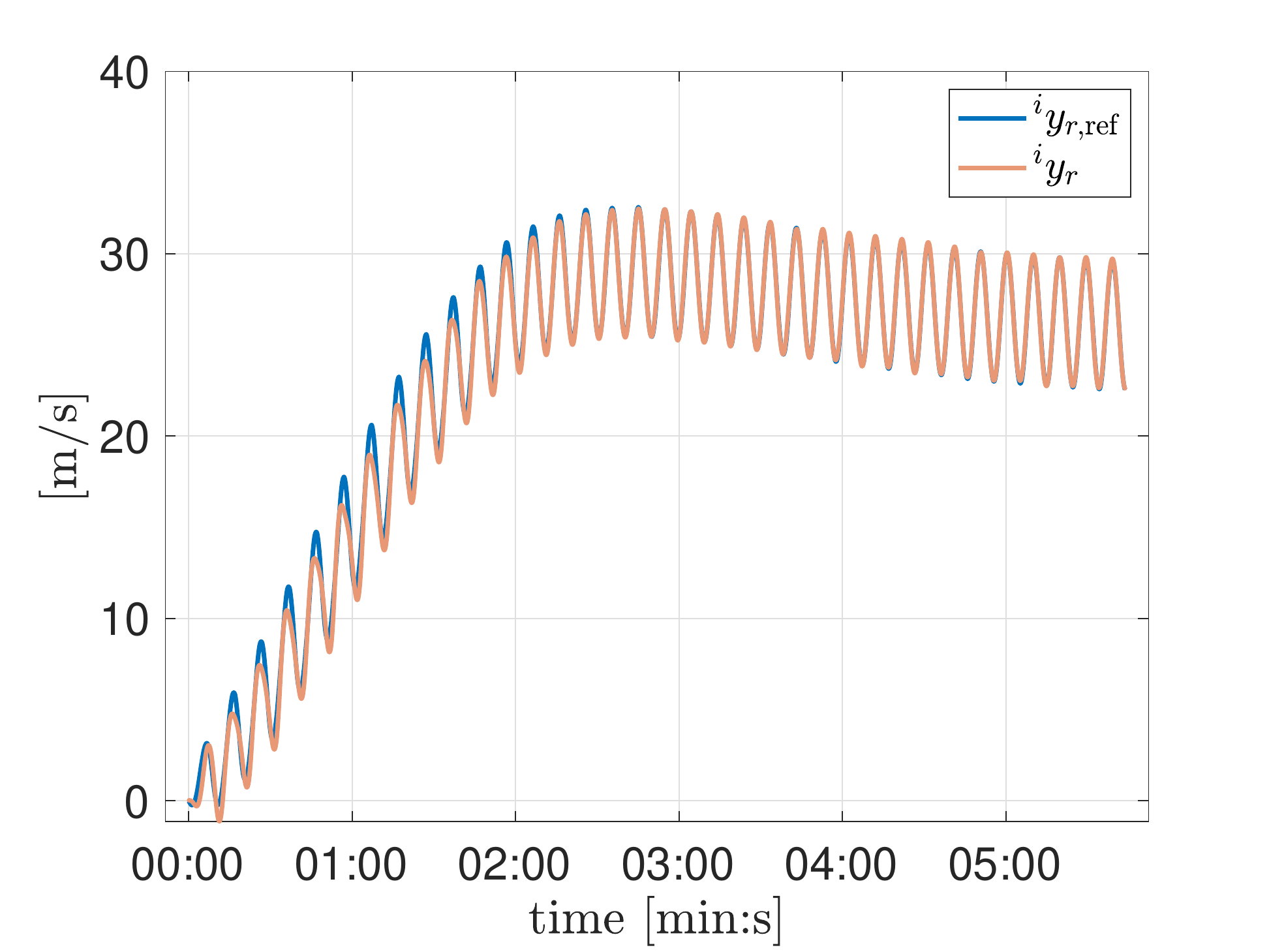}
		\caption{Reference tracking: $y$-component.}
		\label{fig:sim_results(e)}
		\vspace{2pt}
		\centering
		\includegraphics[clip = true, width = \textwidth, height = 0.8\textwidth]{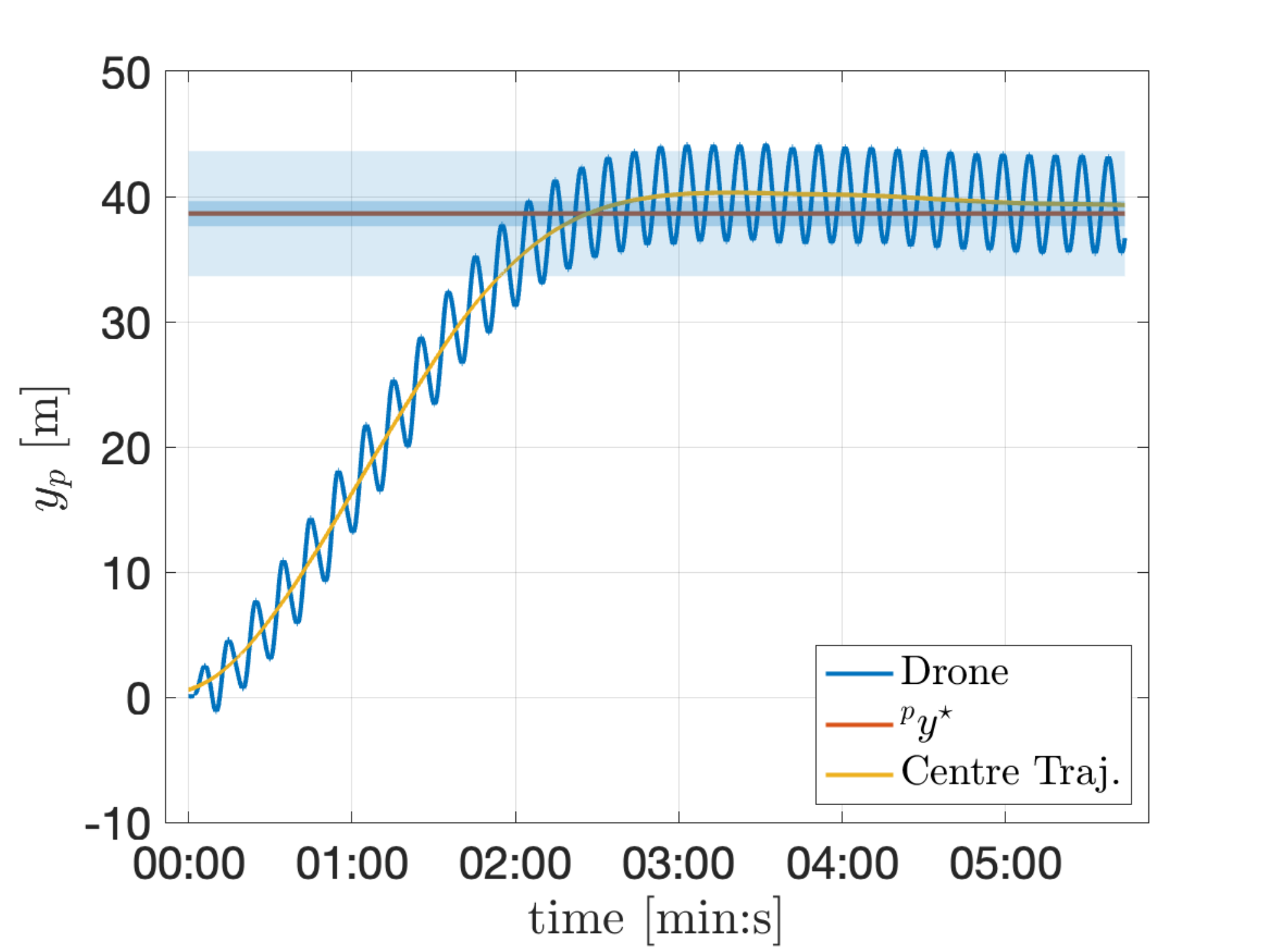}
		\caption{Optimum Seeking: $y$-component.}
		\label{fig:sim_results(f)}
	\end{subfigure}
	\hspace{0.01\textwidth}
	\begin{subfigure}[b]{0.31\textwidth}
		\centering
		\includegraphics[clip = true, width = \textwidth, height = 0.8\textwidth]{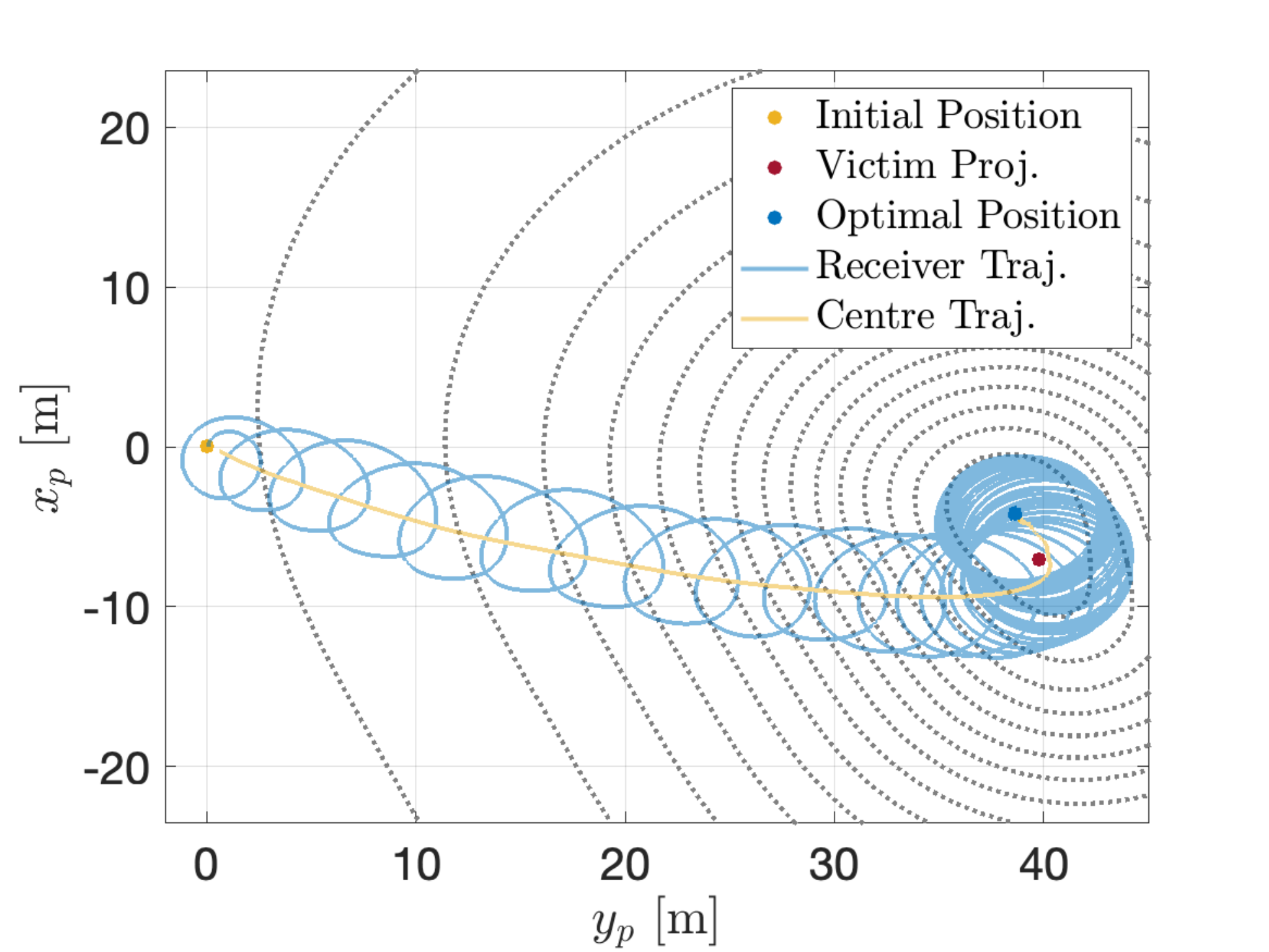}
		\caption{Drone trajectory: $x_py_p$-plane view.}
		\label{fig:sim_results(g)}
		\vspace{2pt}
		\centering
		\includegraphics[clip = true, width = \textwidth, height = 0.8\textwidth]{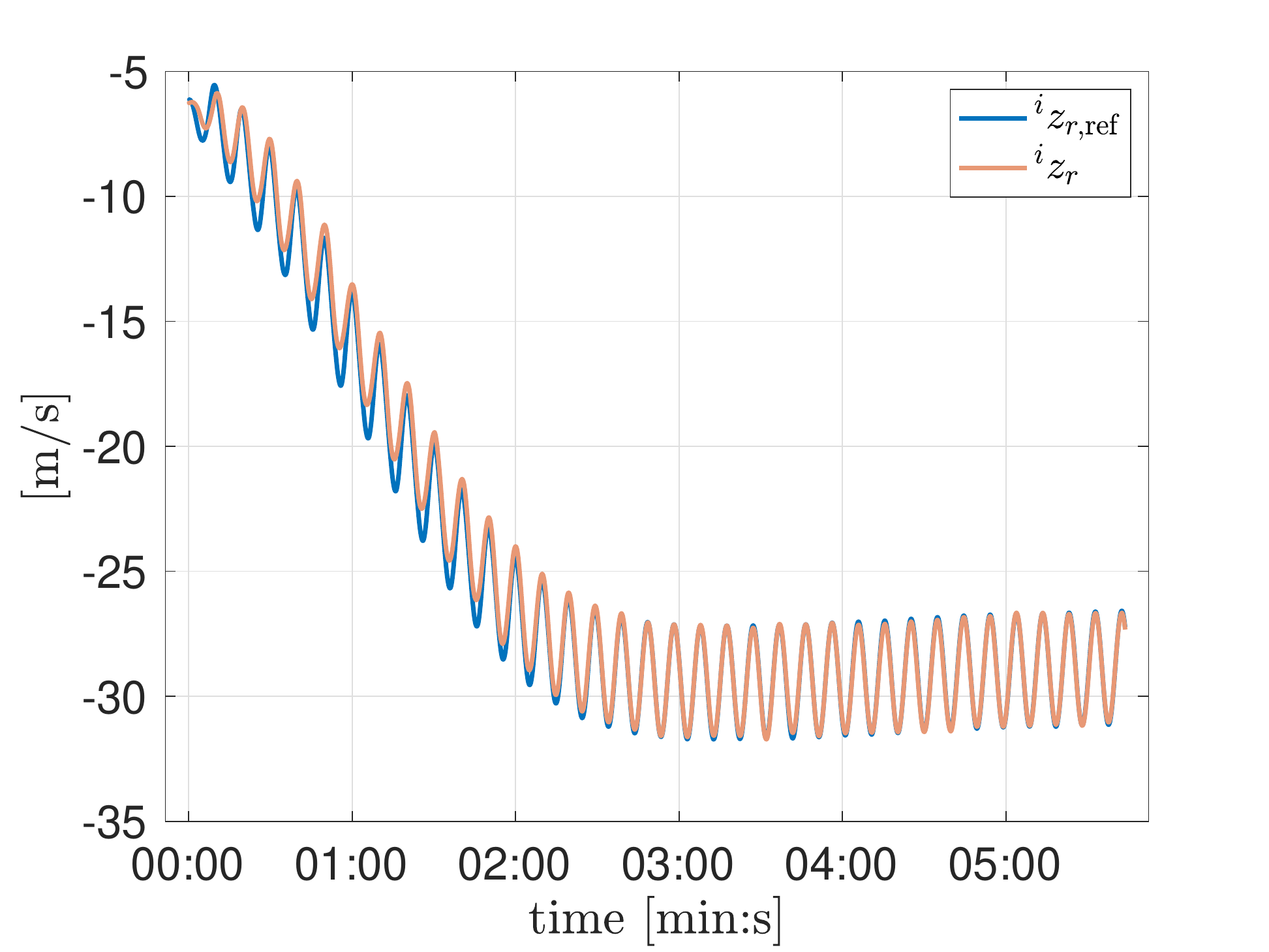}
		\caption{Reference tracking: $z$-component.}
		\label{fig:sim_results(h)}
		\vspace{2pt}
		\centering
		\includegraphics[clip = true, width = \textwidth, height = 0.8\textwidth]{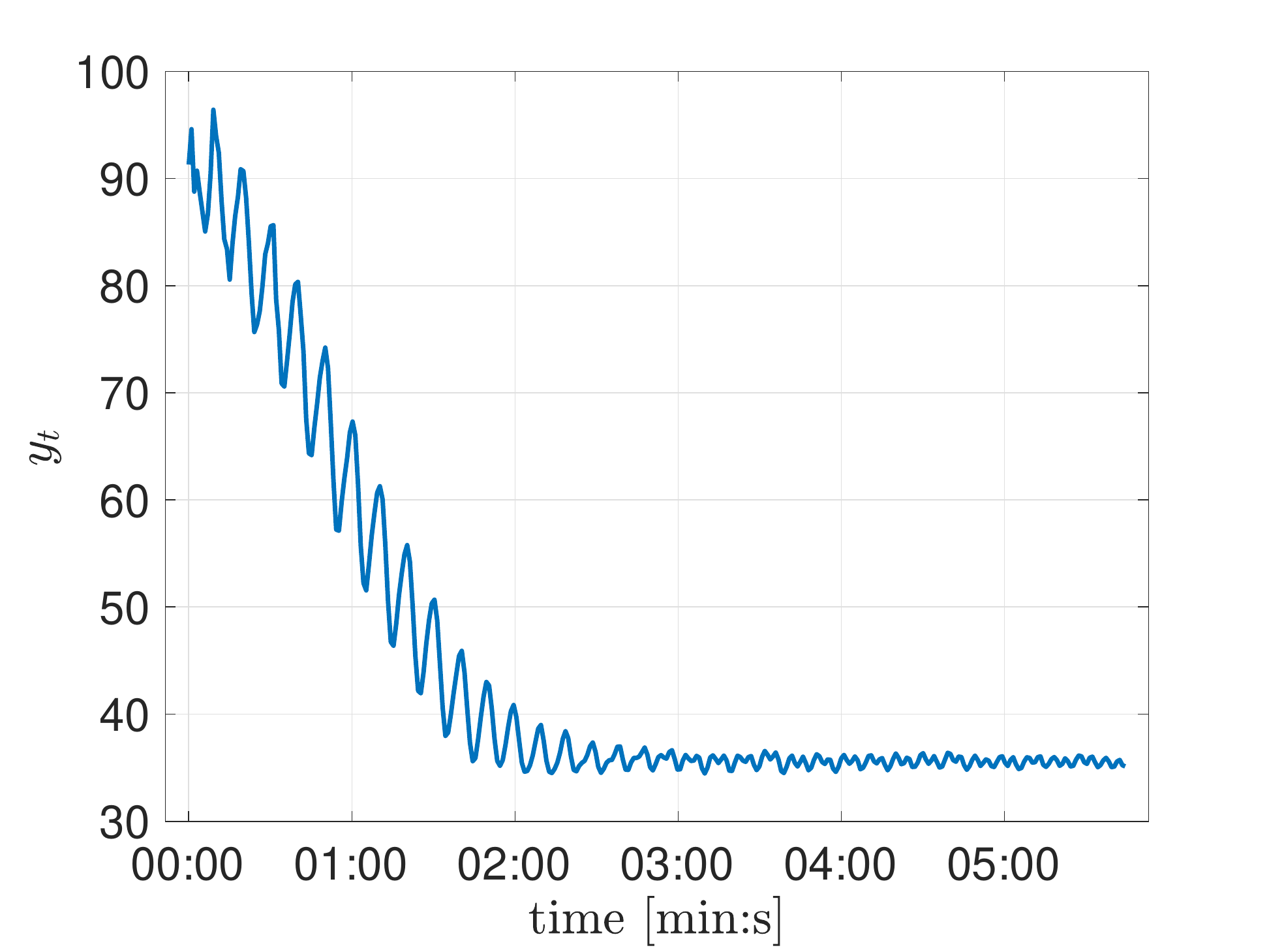}
		\caption{Conditioned ARVA signal.}
		\label{fig:sim_results(i)}
	\end{subfigure}
	\caption{Results of the SITL simulation.}
	\label{fig:sim_results}
	\vspace{-5pt}
\end{figure*} 
\begin{figure*}[t!]
	\centering
	\begin{subfigure}[b]{0.31\textwidth}	
		\centering
		\includegraphics[clip = true, width = \textwidth, height = 0.8\textwidth]{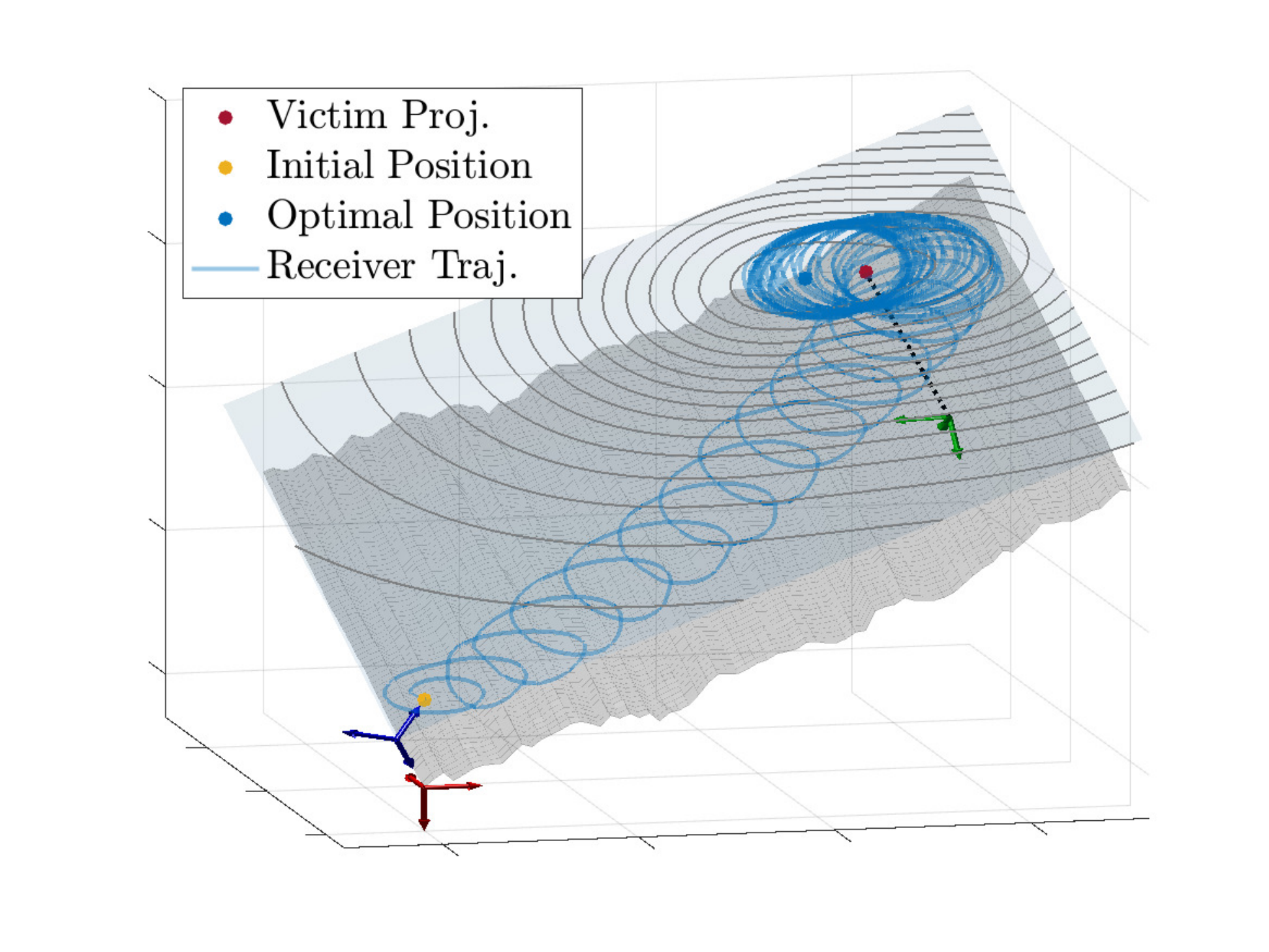}
		\caption{Drone trajectory: 3D view.}
		\label{fig:sim_results_hw(a)}
		\vspace{2pt}	
		\centering
		\includegraphics[clip = true, width = \textwidth, height = 0.8\textwidth]{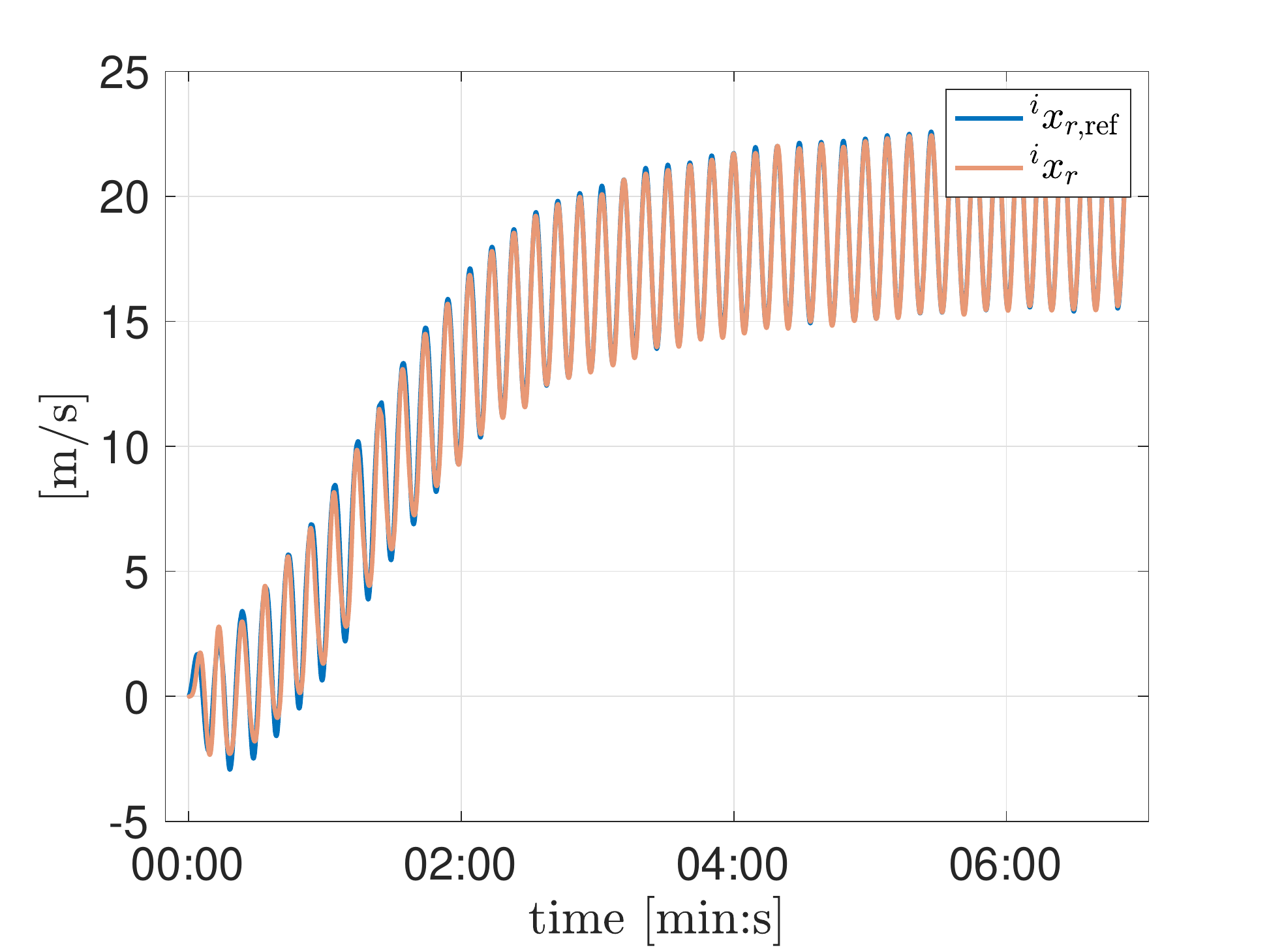}
		\caption{Reference tracking: $x$-component.}
		\label{fig:sim_results_hw(b)}
		\vspace{2pt}
		\centering
		\includegraphics[clip = true, width = \textwidth, height = 0.8\textwidth]{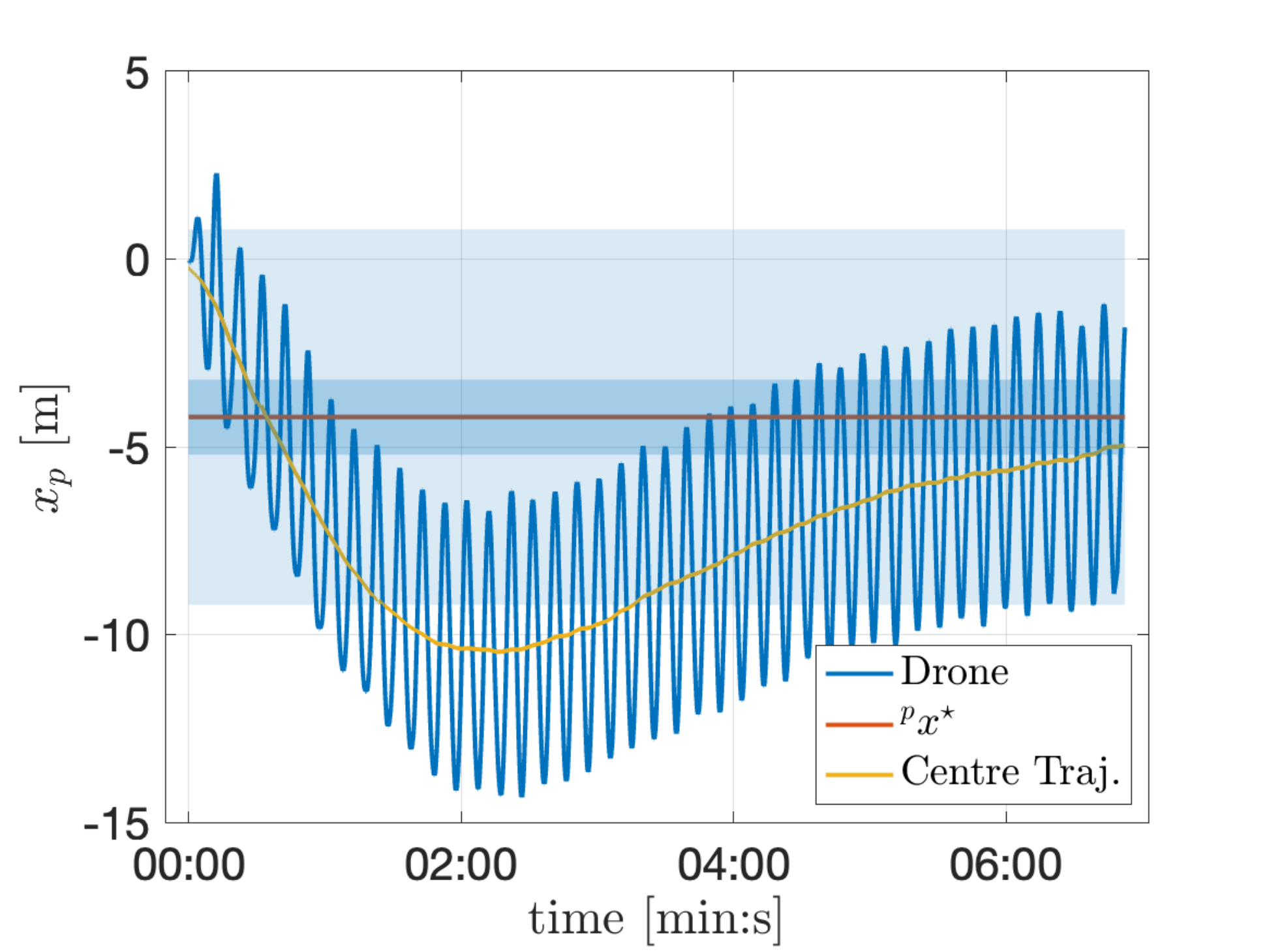}
		\caption{Optimum seeking: $x$-component.}
		\label{fig:sim_results_hw(c)}
	\end{subfigure}
	\hspace{0.01\textwidth}
	\begin{subfigure}[b]{0.31\textwidth}
		\centering
		\includegraphics[clip = true, width = \textwidth, height = 0.8\textwidth]{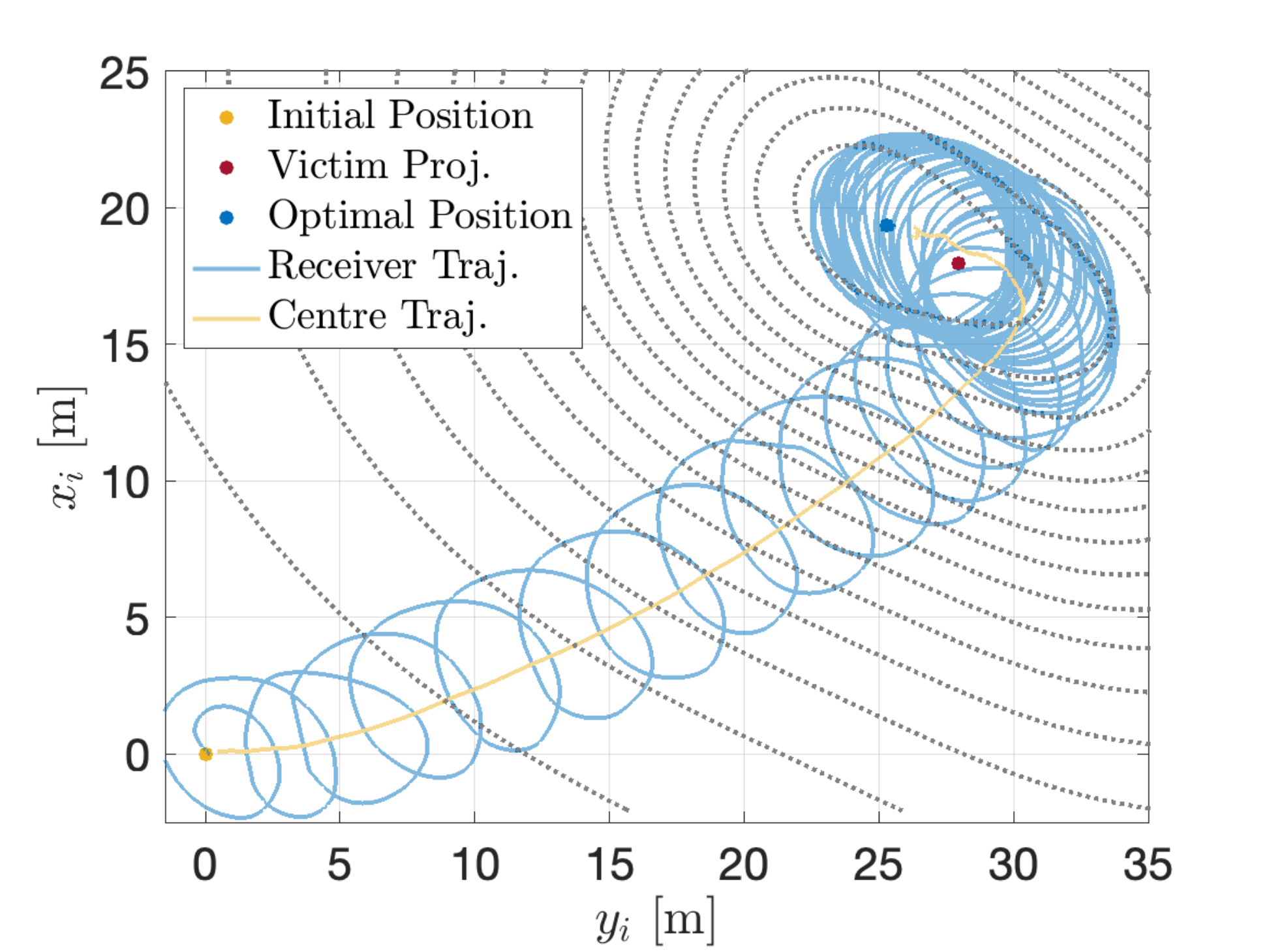}
		\caption{Drone trajectory: $x_iy_i$-plane view.}
		\label{fig:sim_results_hw(d)}
		\vspace{2pt}
		\centering
		\includegraphics[clip = true, width = \textwidth, height = 0.8\textwidth]{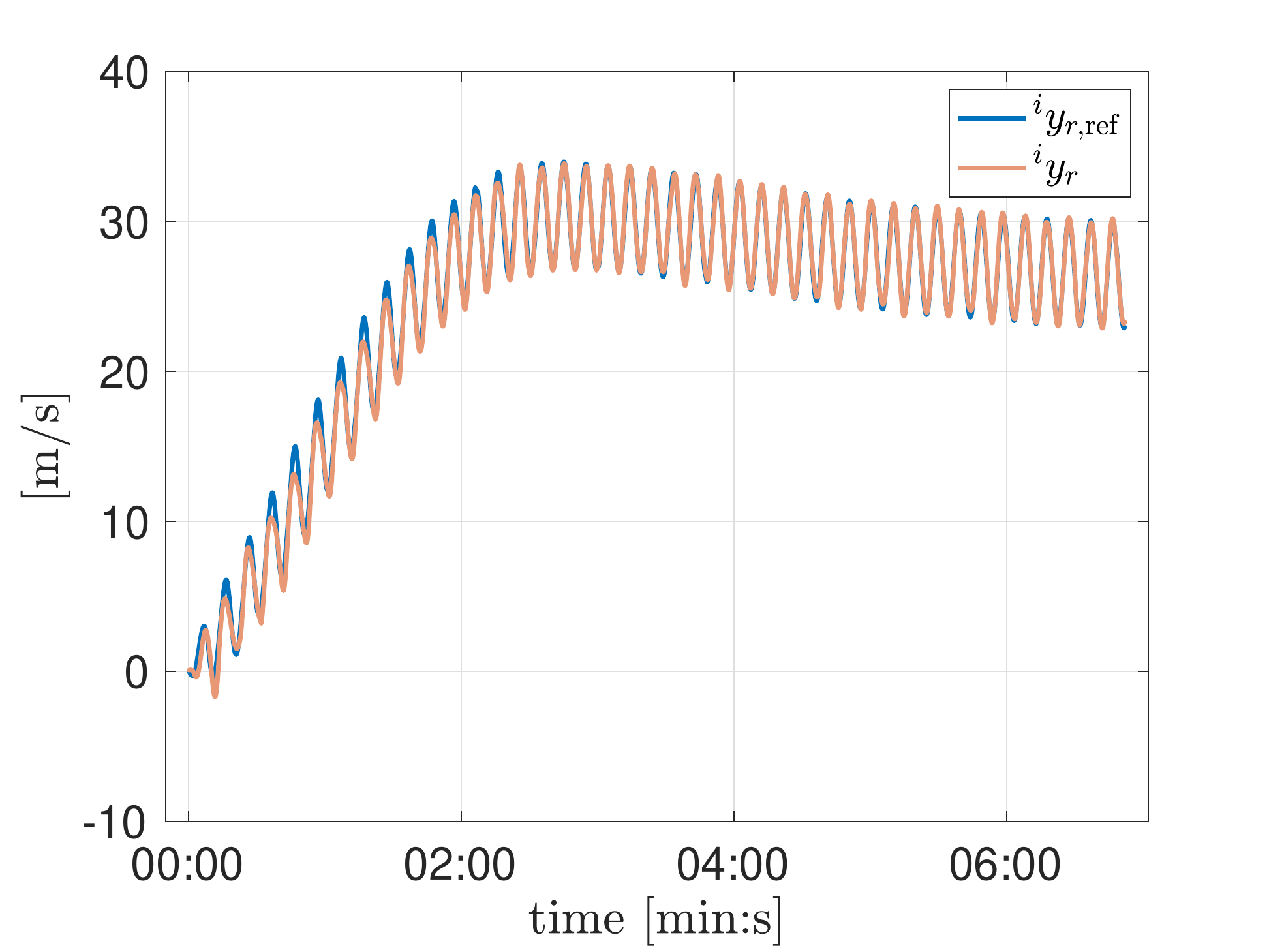}
		\caption{Reference tracking: $y$-component.}
		\label{fig:sim_results_hw(e)}
		\vspace{2pt}
		\centering
		\includegraphics[clip = true, width = \textwidth, height = 0.8\textwidth]{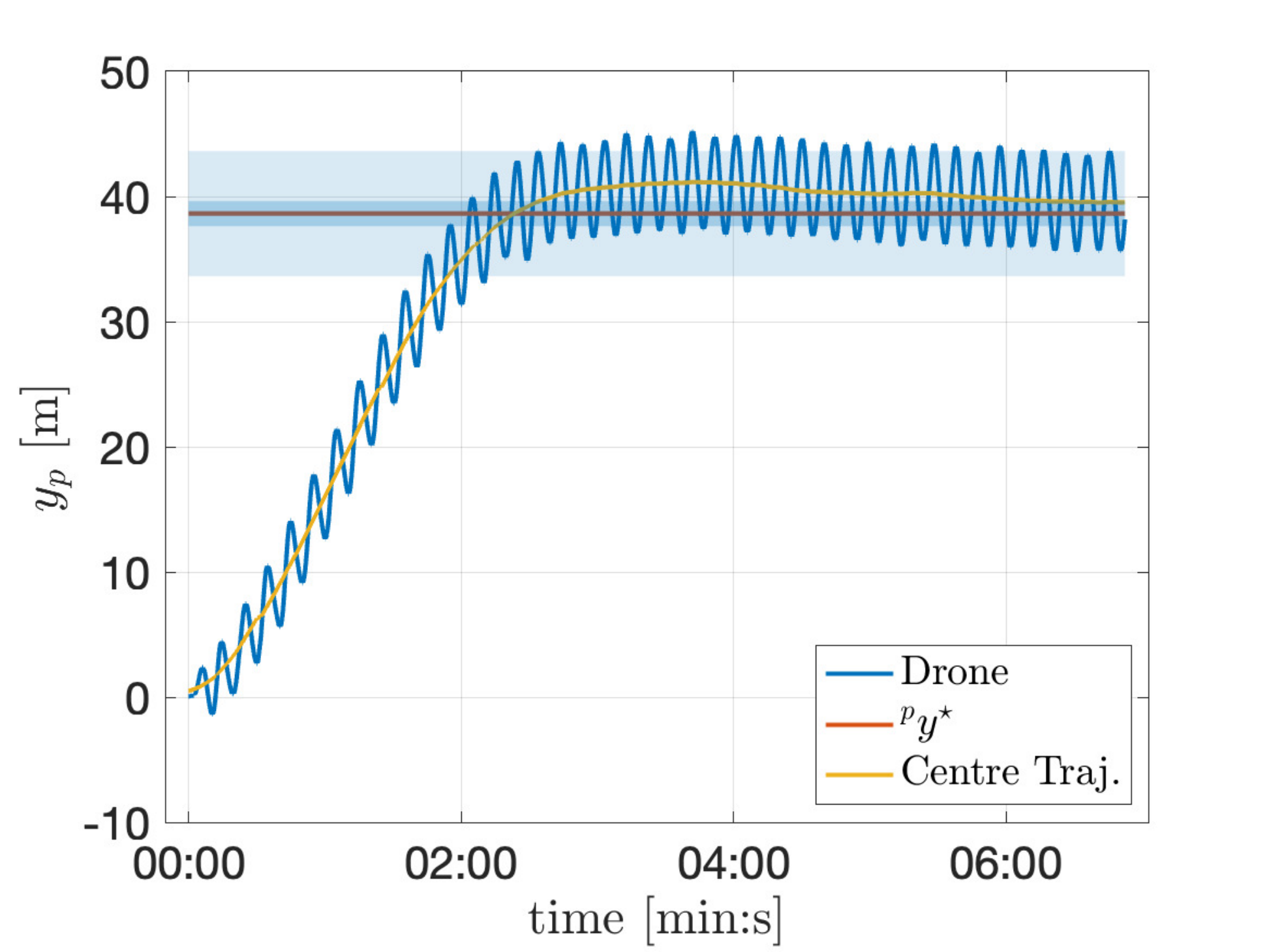}
		\caption{Optimum seeking: $y$-component.}
		\label{fig:sim_results_hw(f)}
	\end{subfigure}
	\hspace{0.01\textwidth}
	\begin{subfigure}[b]{0.31\textwidth}
		\centering
		\includegraphics[clip = true, width = \textwidth, height = 0.8\textwidth]{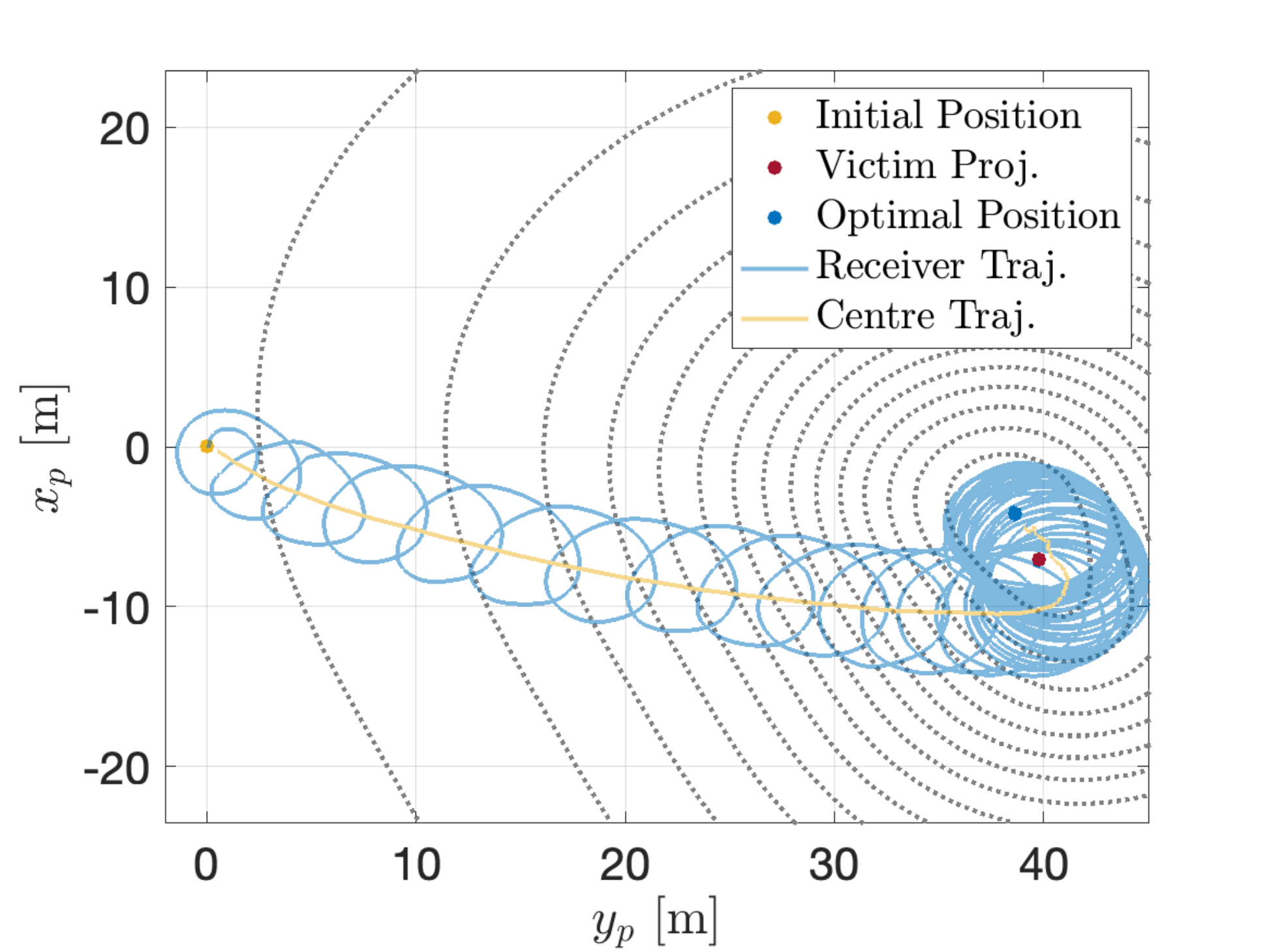}
		\caption{Drone trajectory: $x_py_p$-plane view.}
		\label{fig:sim_results_hw(g)}
		\vspace{2pt}
		\centering
		\includegraphics[clip = true, width = \textwidth, height = 0.8\textwidth]{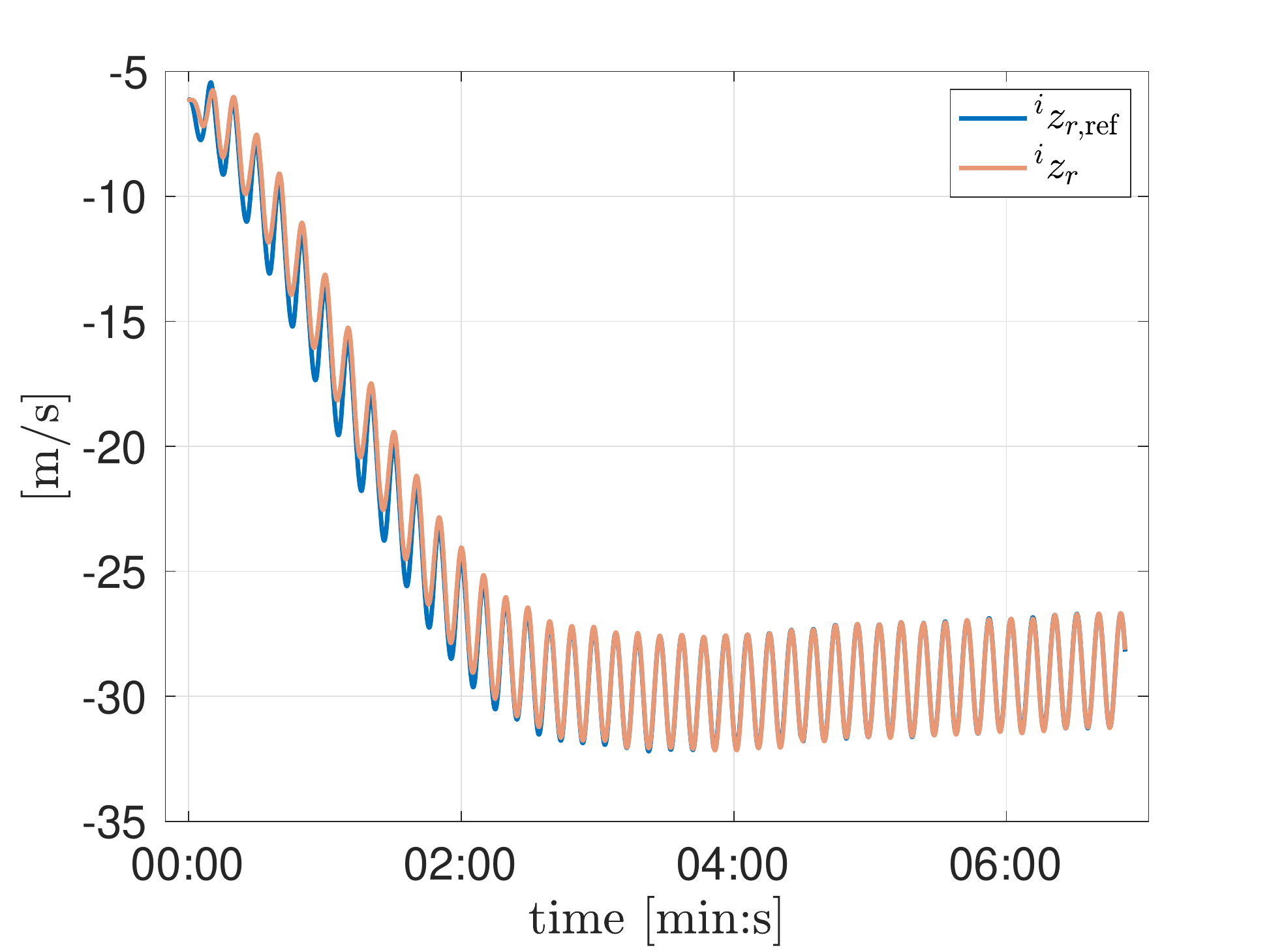}
		\caption{Reference tracking: $z$-component.}
		\label{fig:sim_results_hw(h)}
		\vspace{2pt}
		\centering
		\includegraphics[clip = true, width = \textwidth, height = 0.8\textwidth]{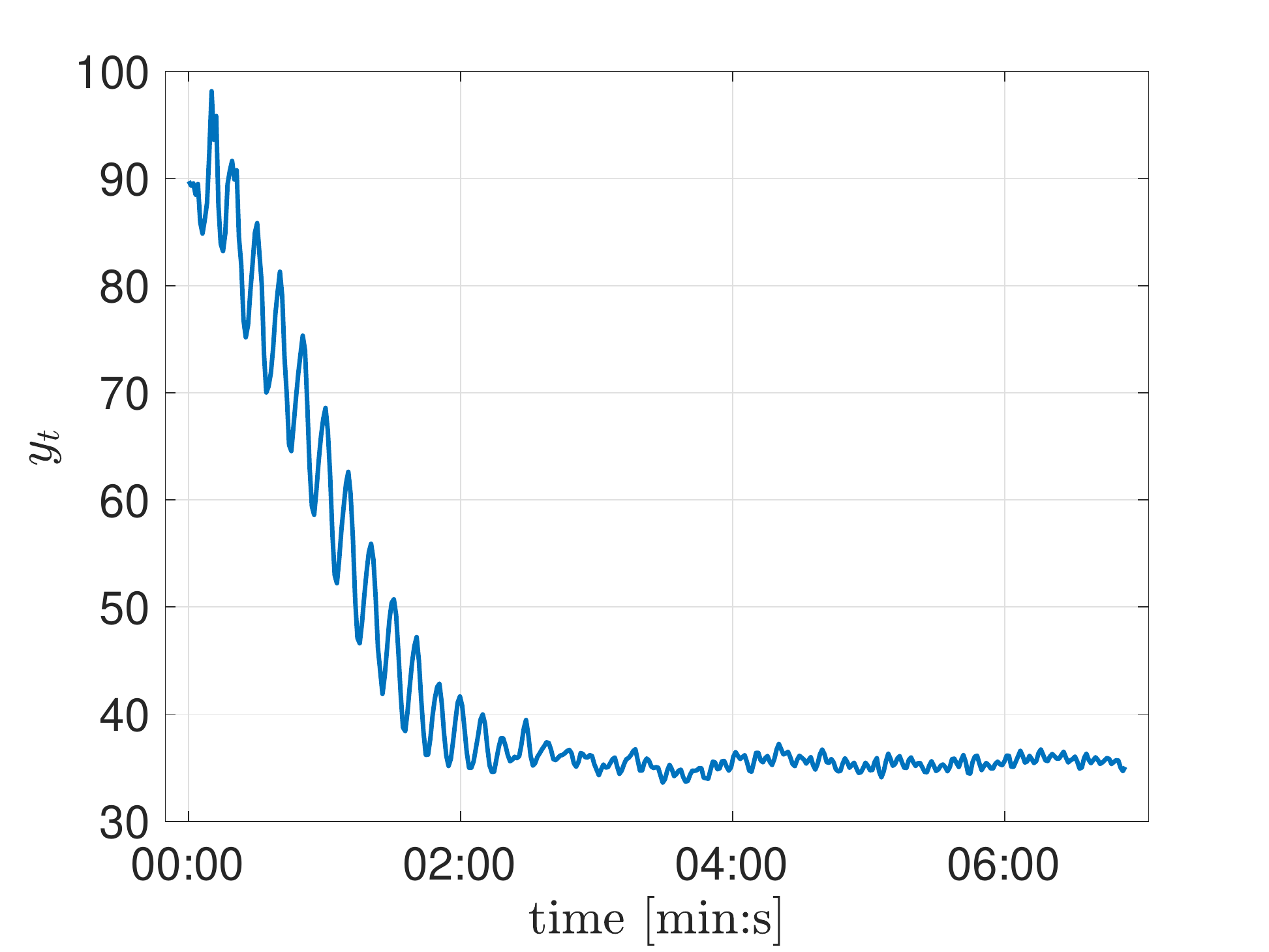}
		\caption{Conditioned ARVA signal.}
		\label{fig:sim_results_hw(i)}
	\end{subfigure}
	\caption{Results of the HITL simulation.}
	\label{fig:sim_results_hw}
	\vspace{-5pt}
\end{figure*} 

To get closer to real use cases, the Gazebo simulation environment has been shaped to mimic an avalanche scenario, where the drone cannot flight at a fixed altitude with respect to the inertial frame due to the mountain slope, unlike our previous work \cite{Azzollini2020Extremum}.
The homogeneous transformation from the search plane to the inertial reference frame $^iH_p$ is defined as:
\begin{equation*}
	^iH_p = \left[\begin{array}{cc}
		^iR_p & ^iO_p \\ 0 & 1
	\end{array}\right].
\end{equation*}
The chosen simulation parameters are summarized in Table \ref{tab:SimParam}

The proposed algorithm was discretized as follows. The discrete-time low-level controller was obtained by: (i) using the zero-order-hold for discretizing the four plants; (ii) the Tustin's method for discretizing the internal model units; (iii) discrete-time Linear Quadratic Regulation theory to obtain the control gains, so as to force the drone to be in near hovering conditions at all times. The ES algorithm was discretized by means of the simple forward Euler method, while the $\alpha$-filter was discretized using Tustin's method once again.
The low-level controller is designed to work at a fixed frequency of $250 \  \rm Hz$, the ES algorithm runs at $10 \  \rm Hz$ while the frequency of the ARVA signals is the lowest, being $1 \  \rm Hz$.

The ES parameters have been carefully designed following the rationale presented at the end of Section \ref{sec:Esunit}, taking into consideration a maximum feasible velocity of $4 \ \rm m/s$ as platform architectural limit. In particular, the best results have been obtained setting $\alpha = 20$, $\kappa = 0.07$ and $\omega = 0.65$, that correspond to a steady-state radius of $3.6$ meters and a maximum drone velocity (on the search plane) of $3.6 \  \rm m/s$.
Moreover, a low-pass filter has been implemented in order to damp the noise affecting the ARVA signal.

\subsection{SITL Simulations}
SITL simulations have been performed on a PC running Ubuntu 18.04.3 LTS with Intel(R) Core i7-3770K@3.60 GHz CPU and 32 GB RAM. The main results are reported in \figref{fig:sim_results}. In particular, \figref{fig:sim_results(c)} and \figref{fig:sim_results(f)} compare the drone trajectories on the search plane (blue line), with both the trajectory of the circumference centre (yellow line) and the position of the ARVA minimizer on the search plane (red line). The dark blue shaded area in \figref{fig:sim_results(c)} and \figref{fig:sim_results(f)} represents a bounding box of dimensions $1\times1$ meters around the optimum position, such interval has been chosen as the interval of practical convergence.
On the other hand, the light blue shaded area draws a bounding box of dimension $5\times5$ meters around the optimum, this interval represents the minimum distance required from experienced rescues in order to find a buried victim. From these figures, it is clear that the practical convergence to the optimum is obtained in approximately $300$ seconds, with the loitering circumference centre entering inside the bigger bounding box before $150$ seconds.
In \figref{fig:sim_results(a)} is reported the 3-dimensional drone trajectory (dark blue) on the search plane (light blue) and the iso-power lines of the ARVA function along the search plane (in gray), while in \figref{fig:sim_results(d)} and \figref{fig:sim_results(g)} the same quantities are projected on the $x_iy_i$-plane and on the search plane, respectively, for better visualization. From the aforementioned figures it is possible to see how the ES algorithm steers the drone-receiver towards the optimum by performing, as expected, a circlular trajectory whose centre follows an approximate gradient descent direction. 
This behavior is particularly visible in \figref{fig:sim_results(d)} and \figref{fig:sim_results(g)}, where the yellow line represents the trajectory of the loitering circumference centre.
Recall that the sought minimum does not coincide with the projection on the search plane of the victim position. This is visible in \figref{fig:sim_results(a)}, \figref{fig:sim_results(d)}, and \figref{fig:sim_results(g)}, where the red dot represents the victim position projected on the search plane $p_{t / {\rm proj}}$, while the blue one is the optimal position (ARVA minimizer) on the same plane $p^\star$. Finally, \figref{fig:sim_results(i)} shows the behavior of the ARVA signal, while \figref{fig:sim_results(b)}, \figref{fig:sim_results(e)}, and \figref{fig:sim_results(h)} report the true drone-receiver inertial positions versus the requested ones. Note that, the non-negligible motion of the circles centre, during the transient, causes a mismatch between the adopted internal model and the model of the reference trajectory, leading inevitably to non-zero tracking errors.
However, notice that the low-level controller manages to keep the tracking errors very small at all times, thus ensuring the needed time scale separation. Moreover, at steady-state, when the reference signals truly become simply biased sinusoids, the tracking errors are practically zero, thus resulting in a better estimate of the optimum.
\subsection{HITL Simulations}
In order to verify the usability on real applications, the proposed solution has been tested on a low-cost microcontroller with limited capabilities. In particular, HITL simulations have been performed on the \textit{Pixhawk 2 Cube} board,  endowed of a STM32F427 Cortex-M4F(R)@168 MHz (252 MIPS) core, with FPU and 256 KB RAM. The obtained results are reported in \figref{fig:sim_results_hw}, which presents the same images configuration proposed in \figref{fig:sim_results}, so as to facilitate the comparison.
Notice that despite some numerical errors, which induce a degradation of the tracking performance, the practical convergence to the bounding box of $1 \times 1$ meters, is still obtained in approximately $390$ seconds, while the larger bound is broken after only $180$ seconds.
Thanks to its lightweight, the developed algorithm is still able to run at a fixed frequency of $250 \ \rm{Hz}$ jointly with the ES module, running at $10 \ \rm{Hz}$. The source code can be found at \href{https://github.com/casy-lab/PX4\_Firmware}{https://github.com/casy-lab/PX4\_Firmware}.

\section{Conclusions}\label{sec:conclusions}
In this work, we presented a complete control architecture for a UAV which, being equipped with an ARVA receiver, is able to autonomously explore the area of interest and converge as close as possible to the victim-ARVA transmitter.

The scheme presented in Figure \ref{fig:ControlArchitecture} is general, in the sense that it could be used for any source seeking control problem where a mobile robot needs to be driven towards a source. The particular choice of an ES control algorithm, and of a low-level controller, depends both on the specific application and on the technology at our disposal. In general, ES should generate a reference trajectory that is dynamically feasible for the specific mobile robot, while the low-level controller should be designed considering both the specific robot and the chosen ES algorithm.

Finally, we extended the well-known PX4 flight stack by creating a new flight mode where our ES reference generator as well as our low-level controller are used. The proposed algorithm performs well even in HITL simulations, converging in a reasonably good amount of time, proving robustness with respect to noise, and providing a very good estimate of the projection of the victim position on the search plane, and thus also on the snow/terrain plane (which is the optimal point from which digging should be performed). The code is available open source to encourage usage as well as possible external contributions.

\bibliographystyle{unsrt}

\begin{IEEEbiography}[{\includegraphics[width=1in,height=1.25in,clip,keepaspectratio]{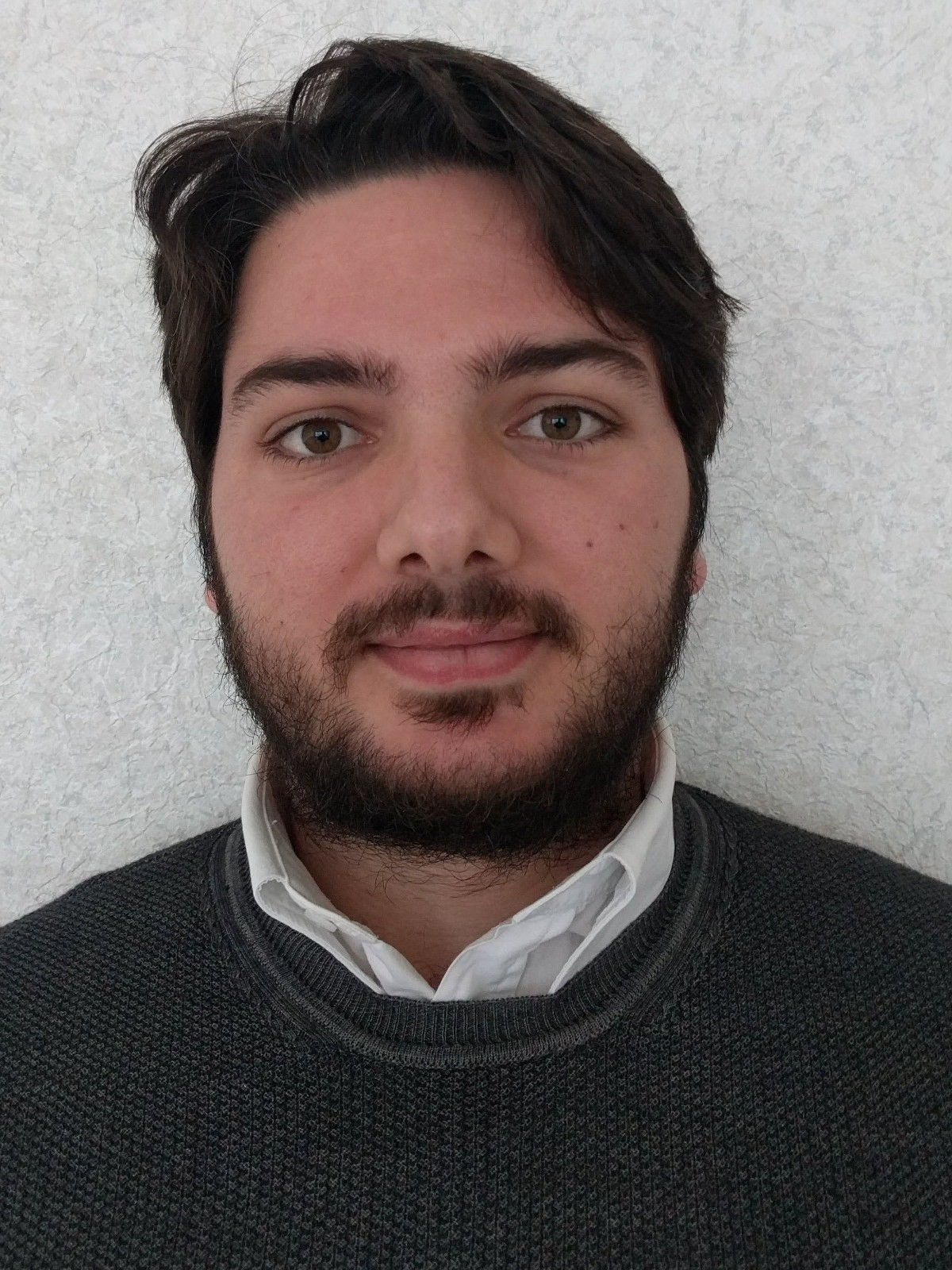}}]{Ilario Antonio Azzollini} received the B.Sc. degree in Automation Engineering from Alma Mater Studiorum University of Bologna, Italy, 2016, and the MSc degree Cum Laude in Systems and Control from the Delft University of Technology, The Netherlands, 2018. He is currently pursuing the Ph.D. degree in Automatica at University of Bologna. His research interests include nonlinear optimization, nonlinear and adaptive control, and control of mobile robots.
\end{IEEEbiography}

\begin{IEEEbiography}[{\includegraphics[width=1in,height=1.25in,clip,keepaspectratio]{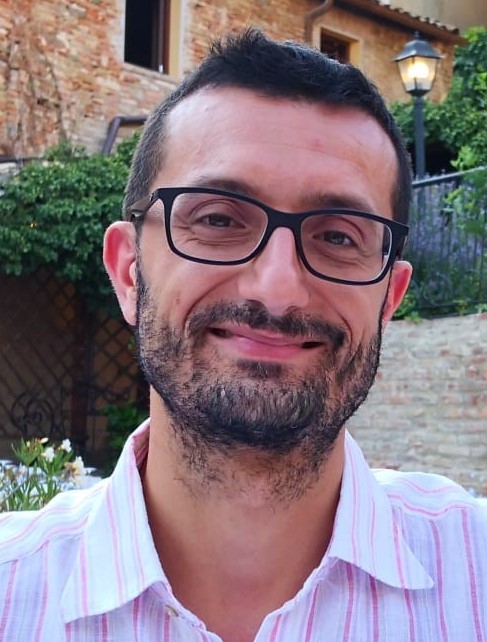}}]{Nicola Mimmo} received the Ph.D. in automation from University of Bologna, in 2015. Since 2010, he has been working with the Department of Electrical, Electronic, and Information Engineering ``Guglielmo Marconi'', University of Bologna, on national and European projects for the development of unmanned aerial vehicles for civil applications with focus on search and rescue scenarios. He worked with the major national and European aircraft companies for the development of both flight systems and flight control laws that are currently protected by international patents.  He is co-author of tens of scientific papers. His research interests range from fault-tolerant controls to nonlinear control systems in aerospace.
\end{IEEEbiography}

\begin{IEEEbiography}[{\includegraphics[width=1in,height=1.25in,clip,keepaspectratio]{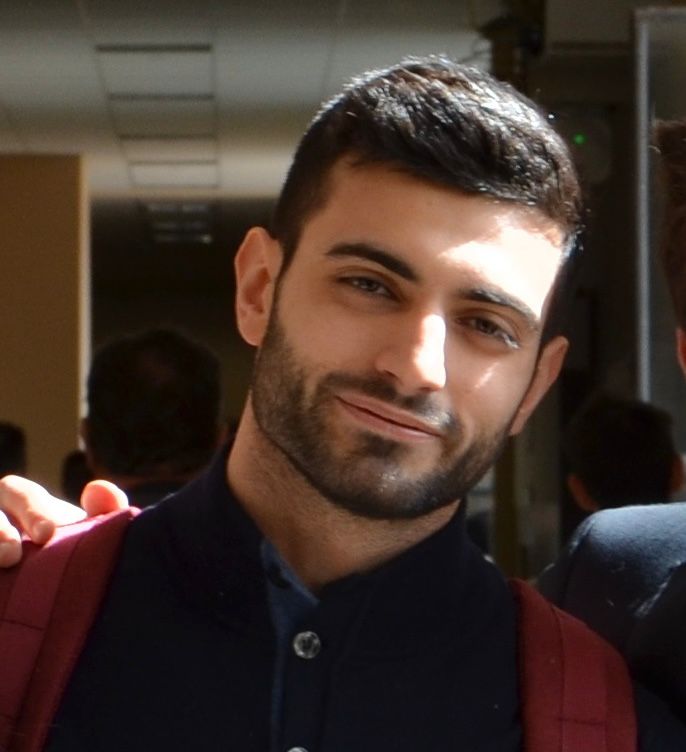}}]{Lorenzo Gentilini} received his M.Sc. degree in Automation Engineering from the University of Bologna, Italy, in $2019$. Since $2019$ he works as Ph.D. at University of Bologna. His work focuses on autonomous navigation of medium-size unmanned aerial vehicles, with special attention to motion planning and localization.
His main research interests include trajectory planning algorithms, localization and mapping strategies, robot vision and machine learning applications. Since $2019$, he is also a research fellow at the center for research on Complex Automated SYstems (CASY), working on the \textit{Drone Contest} project.
\end{IEEEbiography}

\begin{IEEEbiography}[{\includegraphics[width=1in,height=1.25in,clip,keepaspectratio]{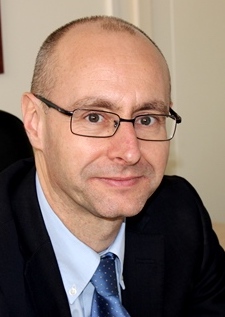}}]{Lorenzo Marconi} (Fellow, IEEE) graduated in 1995 in Electrical Engineering from the University of Bologna. Since 1995 he has been with the Department of Electronics, Computer Science and Systems at University of Bologna, where he obtained his Ph.D. degree in March 1998. From 1999 he has been an Assistant Professor in the same Department where is now Full Professor since January 2016.
 
He has held visiting positions at and collaborations with various academic/research international institutions. He is co-author of more than 250 technical publications on the subject of linear and nonlinear feedback design published on international journals, books and conference proceedings. He is also co-author of three international monographs.
In 2005, he has been awarded jointly by Elsevier and the International Federation of Automatic Control (IFAC) for the best paper published in the period 2002-2005 on ``Automatica''. He is also the recipient of the 2014 IEEE Control Systems Magazine Outstanding Paper Award for the best paper published on the magazine in the period 2012-2013, He is the recipient of the 2018 O. Hugo Schuck Best Paper Award assigned by the American Automatic Control Council for the best paper presented at the 2017 American Control Conference. He is Fellow of IEEE for ``contributions to feedback design of nonlinear systems and unmanned aerial vehicles''.

He served as associate editors of the main international journals in the field of control, such as Automatica, IEEE Transaction on Automatic Control, and IEEE Control Systems Technology. He is now serving as Senior Editor of IEEE Transaction on Automatic Control.
 
His current research interests include nonlinear control, output regulation and stabilisation of nonlinear systems, control of autonomous aerial vehicles, robust control, fault detection and isolation, fault tolerant control.
\end{IEEEbiography}

\end{document}